\documentclass[10pt,twocolumn,letterpaper]{article}
\usepackage[pagenumbers]{cvpr}

\definecolor{cvprblue}{rgb}{0.21,0.49,0.74}
\usepackage[pagebackref,breaklinks,colorlinks,allcolors=cvprblue]{hyperref}
\def\confName{CVPR}
\def\confYear{2027}

\usepackage{uoftcolors}
\usepackage{graphicx}
\usepackage{subcaption} 
\usepackage{placeins}

\usepackage{pifont}
\newcommand{\cmark}{\textcolor{uoftdarkgreen}{\ding{51}}}
\newcommand{\xmark}{\textcolor{uoftwarmred!70!black}{\ding{55}}}

\usepackage{etoolbox}
\patchcmd{\abstract}
{\vspace*{12pt}}
{\vspace*{4pt}}
{}
{\PackageWarning{elmo}{Abstract spacing patch failed}}

\title{ELMO: Emergent Representations In Imitation Learning \\ Via Dynamic Hierarchical Action Chunking}

\author{
    Jasper Gerigk$^{*1,2}$ \quad
    Yang Xu$^{*1,2}$ \quad
    Nicholas Rhinehart$^{\dagger 1}$ \quad
    Igor Gilitschenski$^{\dagger 1,2}$ \\[0.5em]
    $^1$University of Toronto \quad
    $^2$Vector Institute \\[0.3em]
    {\small $^*$Equal contribution \quad
    $^\dagger$Equal supervision}
}

\begin{document}

\twocolumn[{%
  \renewcommand\twocolumn[1][]{#1}%
  \maketitle
  \begin{center}
    \centering
    \captionsetup{type=figure}
    \vspace{-0.8cm}
    \includegraphics[trim={0cm 0.1cm 0cm 0.1cm},clip,width=1.0\textwidth]{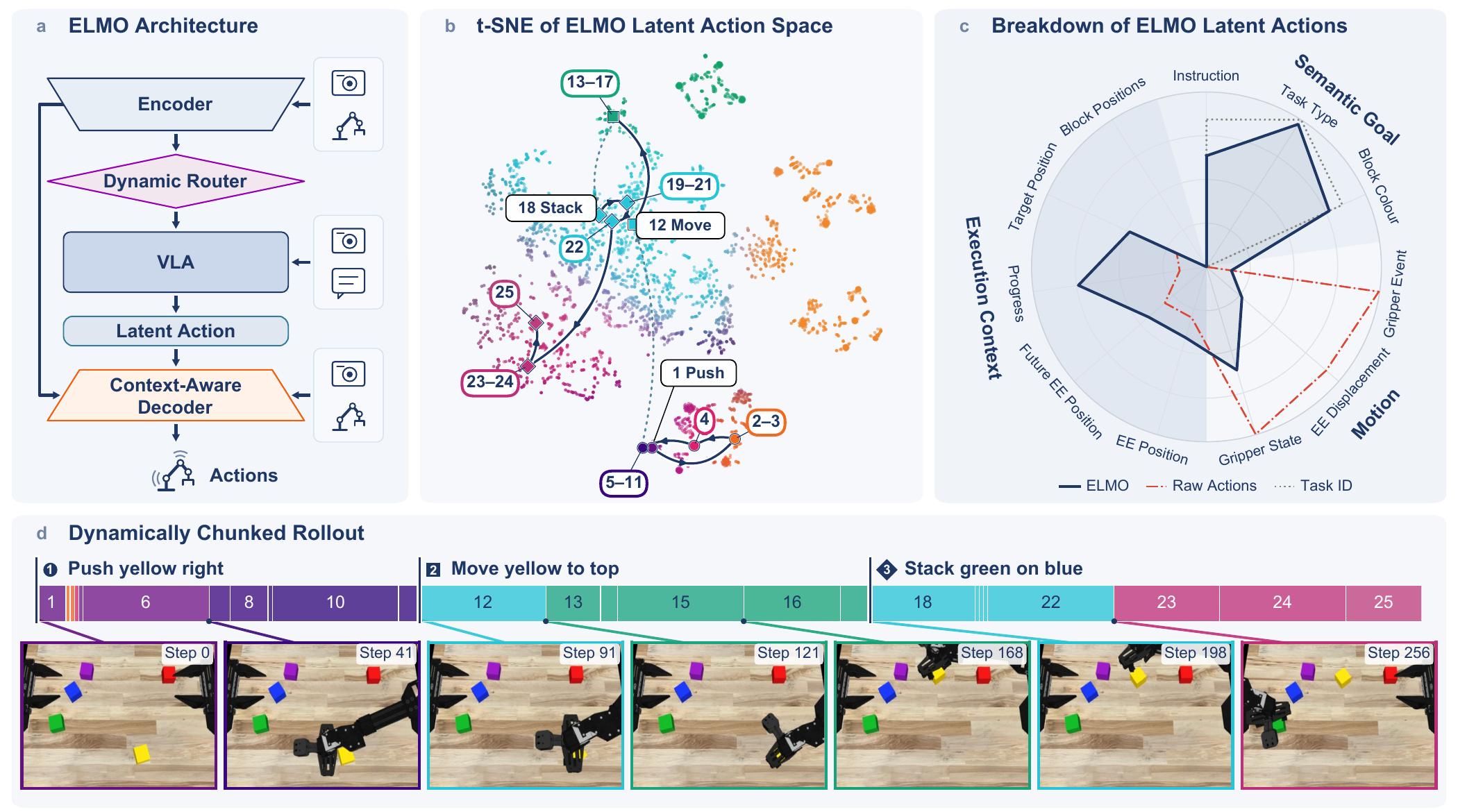}
    \captionof{figure}{ELMO is a jointly-trained hierarchical policy (a) that learns semantically grounded latent actions (b,c) and splits trajectories into segments by shared intent (d). Chunk latents from (d) are traced in (b), with similar colours indicating proximity in latent space.}
    \label{fig:teaser}
  \end{center}
}]

\begin{abstract}
Robot trajectories naturally decompose into variable-length phases with distinct semantic intents, such as reaching, grasping, and lifting.
However, training policies to minimize MSE treats all actions as equal, obscuring their semantic structure.
While latent action models attempt to create a structured prediction target by compressing action chunks into latent representations, existing approaches rely on two-stage training and context-blind decoders.
Consequently, their representations are context-agnostic \textit{motion codes} (e.g., ``move down'') which fail to capture behavioural intent (e.g., ``pick up the block'').
To overcome these bottlenecks, we introduce the \textit{Emergent Latent-action MOdel}~(ELMO), an end-to-end framework that encodes robot trajectories as \textit{grounded semantic codes}---representations that ground task-level behavioural intent in the current environment.
ELMO jointly trains a VLA planner and an observation-conditioned decoder over dynamically chunked trajectories, learning a semantically grounded action space without separate pretraining.
Across standard benchmarks, ELMO outperforms comparable prior methods by 9.6 percentage points on LIBERO-Plus and by +0.75 average task length on CALVIN.
Furthermore, on \textit{Blocks}---our shortcut-free simulation and real-world generalization benchmark---ELMO reaches a 38.6\% success rate while all baselines remain below 5\%, demonstrating its ability to capture the semantic structure of demonstrations. Website: \href{https://uoft-isl.github.io/elmo.github.io/}{https://uoft-isl.github.io/elmo.github.io/}
\end{abstract}

\vspace{-2em}
\section{Introduction}
\label{sec:intro}
Vision-language-action (VLA) policies combine visual and language understanding with robot control by using pretrained vision-language models (VLMs)~\cite{kim2024openvla}.
While the concept of using a VLM backbone is well established, it is unclear how to best parametrise the action prediction task on top of it.
Designs have tried predicting actions via discrete tokens~\cite{kim2024openvla,pertsch2025fast}, diffusion~\cite{chi2025diffusion,wen2025diffusionvla}, or flow matching~\cite{black2024pi_0,pmlr-v305-black25a,wang2026qwen}.
Further, it has been shown that making policies predict a chunk of actions rather than a single one improves performance~\cite{zhao2023act}, as these sequences capture the underlying non-Markovian behaviour and reduce compounding errors~\cite{lazzati2026does}. 
Going beyond the format of actions, training a VLM backbone to directly predict raw actions has been shown to hurt performance due to the large modality gap between language and raw actions~\cite{driess2026knowledge}.
Thus, research has explored alternative ways to represent actions~\cite{sutton1999between,shentu2024llms,li2025hamster}. 

At one extreme, hierarchical methods use language sub-tasks~\cite{ahn2022can,belkhale2024rth,shi2025hi} as \emph{semantic codes} to decompose instructions for a downstream controller. 
However, language remains too coarse, leaving excessive execution uncertainty to the action policy that lacks the full context required to keep trajectories synchronised with the original intent. 
At the other extreme, latent action models compress action chunks into continuous or discrete latents~\cite{lee2024vqbet,zheng2024prise,pertsch2025fast}. 
Yet, as we demonstrate in~\Cref{sec:latent_analysis}, these representations act merely as \emph{motion codes}: context-agnostic encodings of kinematics without task-level intent. 
Ideally, an intermediate representation should combine the best of both as \emph{grounded semantic codes}---representations that capture exact behavioural intent and environmental context while allowing an action policy to reactively adapt execution to the live scene.

Why do existing latent action models collapse into motion codes rather than learning grounded semantic codes? We argue that they are held back by two architectural bottlenecks: context-blind decoding and two-stage training. 
First, prior tokenizers and skill models (e.g., VQ-BeT, PRISE, and FAST~\cite{lee2024vqbet,zheng2024prise,pertsch2025fast}) reconstruct action sequences in isolation without an observation-conditioned decoder. 
Because the decoder is blind to the visual scene, the latent code is forced to memorize exact kinematics rather than abstract behavioural intent. 
Second, even when decoders utilize images~\cite{xu2026edar}, prior works adopt a hierarchical two-stage training paradigm: they first learn to compress actions into a fixed latent space in the first stage, and then train a planner to predict those frozen latents in the second step. 
Decoupling action compression from latent prediction prevents the latent space from being shaped by task-level semantics and visual-language context, optimizing the representation for isolated reconstruction rather than grounded planning.

Overcoming these two bottlenecks requires unifying latent prediction and observation-conditioned decoding into a single end-to-end training stage. 
To this end, we propose \textbf{ELMO}, an end-to-end hierarchical prediction framework that directly trains the VLA planner and the observation-conditioned decoder together, adapting the dynamic temporal chunking mechanism of H-Net~\cite{hwang2025hnet} to continuous robot trajectories. 
Instead of relying on fixed temporal windows, ELMO dynamically allocates latent representations across trajectory boundaries and jointly optimizes the entire hierarchy. 
By training end-to-end in a single stage, the representation space is jointly shaped by vision-language planning and reactive visual control, allowing ELMO to learn a semantically grounded action space. 
During deployment, the VLA planner predicts the resulting \emph{grounded semantic codes}, which the observation-conditioned action policy maps to raw controls. 
Grounded semantic codes serve as a compact, low-noise prediction target, improving VLM learning efficiency while preserving the reactivity and precision of receding-horizon control~\cite{zhao2023act,chi2025diffusion}. 

Our contributions are threefold: (1) we introduce ELMO, an end-to-end hierarchical framework that unifies dynamic chunking with single-stage joint training of a VLA planner and an observation-conditioned decoder to learn grounded semantic codes; (2) we provide a comprehensive analysis of prior latent action methods, characterizing differences in their encoded information; and (3) we demonstrate strong performance on standard benchmarks (LIBERO, LIBERO-Plus, CALVIN) alongside superior generalization on Blocks, our specialized benchmark designed to eliminate evaluation shortcuts in existing suites.

\section{Related Work}
\label{sec:related}

\paragraph{Vision-Language-Action Models.}
In recent years, VLA models have been extensively studied for generalist robotics.
These policies leverage the general knowledge of VLMs by passing VLM embeddings of image and language to an action head for robot control~\cite{kim2024openvla,black2024pi_0}.
The action head predicts future robot actions from the current state.
Following this structure, VLAs have been scaled to billions of parameters and trained on ever-increasing robot datasets~\cite{o2024open}.
Despite this scale, VLAs still struggle to generalize to unseen environments, tasks, and robot embodiments. 
Changes in camera viewpoint, initial robot state, or lighting can cause large drops in success rate~\cite{fei2025liberoplus}.
We argue that this is partly a result of robot policies learning to predict high-frequency actions rather than the intent behind them.
This results in brittle learning and poor generalization, motivating better action representations and training objectives.

\begin{figure*}[t!]
    \centering
    \includegraphics[width=0.84\linewidth]{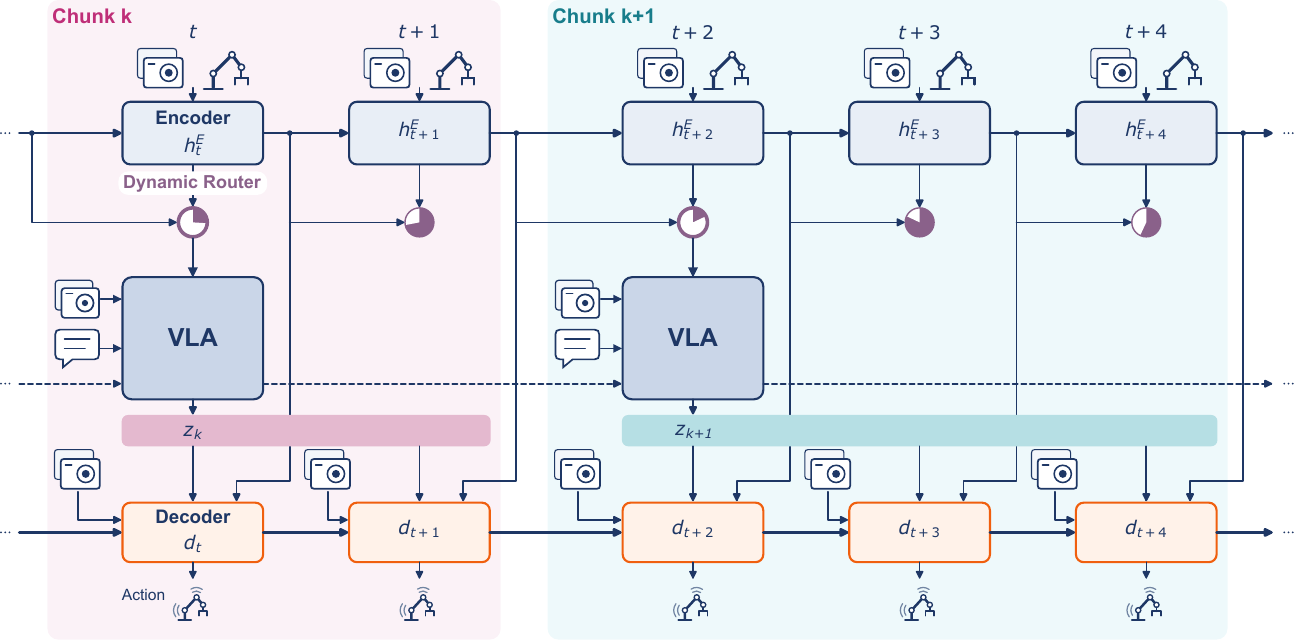}
    \caption{ELMO architecture. At each timestep, the encoder encodes the observations and robot state as $h^E_t$. Comparing it to $h^E_{t-1}$, the router determines whether a chunk should start and the VLA should predict a new latent action using the observations and language. The current latent action along with wrist image and encoder state is used by the recurrent flow-matching decoder to predict actions.}
    \label{fig:method_diagram}
\end{figure*}

\paragraph{Tokenization.}
In vision and language, models generally operate on a tokenized representation of the input rather than raw bytes directly.
By grouping predictable local structures, tokenization shortens sequences and lets large models focus on long-range composition and semantic relationships rather than high-frequency noise.
Vision models typically rely on learned patch embeddings trained end-to-end with the main network. Images are sliced into non-overlapping grid patches, and each flattened patch is mapped through trainable layers into a continuous latent vector~\cite{dosovitskiy2020image}.
Traditionally, language models have relied on separately trained tokenizers like byte-pair encoding (BPE)~\cite{sennrich2016neural}, which starts with a base alphabet and iteratively merges the most frequent adjacent symbol pairs into new tokens.
More recently, learned tokenizers have been developed for language as well.
H-Net~\cite{hwang2025hnet} trains an encoder-backbone-decoder hierarchy end-to-end. Raw byte embeddings pass through an encoder that dynamically groups variable-length spans into learned tokens based on adjacent similarity. The backbone operates in this compressed token space and predictions are decoded back into raw bytes. 
At scales above 1B parameters, H-Net matches the training perplexity and downstream performance of BPE-tokenized transformers while removing the fixed external tokenizer~\cite{hwang2025hnet}.

\paragraph{Action Representations.}
Early VLAs discretized action dimensions into uniform bins to reframe continuous regression as autoregressive classification~\cite{kim2024openvla}, while signal-compression methods like FAST apply the discrete cosine transform and BPE to tokenize short action chunks~\cite{pertsch2025fast}. 
Alternative approaches decompose long-horizon tasks via language sub-goals~\cite{ahn2022can}, finer-grained language instructions~\cite{belkhale2024rth}, or fixed-hierarchy latent codes bridging VLMs and low-level controllers~\cite{shentu2024llms}.
However, these rely on predefined linguistic abstractions and are thus limited in what they can express efficiently. 
Closer to our work, trajectory-level representation methods extract higher-level tokens via vector quantization and sequence compression~\cite{lee2024vqbet,zheng2024prise,yunis2024subwords,mete2024quest,li2025star}, recently scaled up in VQ-VLA~\cite{wang2025vqvla} and systematically benchmarked in ActionCodec~\cite{dong2026actioncodec}. 
Among these, only PRISE constructs variable-length segments via post-quantization BPE, focusing on raw compressibility rather than semantic or temporal structure~\cite{zheng2024prise}. 
Video-based latent action models (e.g., LAPA~\cite{ye2024lapa}, HiLAM~\cite{kim2026hilam}) leverage unlabeled videos and optimize downstream policy integration~\cite{lin2026from}, but learning from visual differences alone risks entangling robot motion with background change~\cite{li2025latbot}. 
Most similar to our method are EDAR and SA-VLA~\cite{xu2026edar,jiang2026savla}. 
Although EDAR incorporates camera observations and predicts visual changes in its decoder, we find it still yields motion codes.
Similarly, SA-VLA conditions its fixed-length decoder on proprioceptive state, but without any information about the environment, it cannot encode anything besides a motion code~\cite{jiang2026savla}. 
More broadly, when decoders lack or underutilize live visual observations, latent tokens must independently reconstruct exact trajectories, collapsing into direct action reparameterizations. 
In contrast, we train a dynamic latent hierarchy end-to-end with a vision-aware decoder, which allows a latent representation carrying task-level intent along with its relevant context to emerge while delegating local execution to the decoder.

\section{Methodology}
\label{sec:methodology}

We build \textbf{ELMO}~(\Cref{fig:method_diagram}) around our two core principles: dynamic chunking of the robot trajectory and a context-aware decoder. 
Since the trajectory chunks should reflect when the policy should make decisions, we train the hierarchical VLA planner and observation-conditioned action policy together. 
ELMO adapts the dynamic chunking mechanism of H-Net~\cite{hwang2025hnet} using a recurrent state encoder and a dynamic router to partition continuous robot trajectories into variable-length behavioural segments.
Training all components together end-to-end via flow-matching action prediction and chunking losses allows us to learn \emph{grounded semantic codes}.
The structure emerges naturally during training and contains information about the current semantic task and execution context (\Cref{sec:latent_analysis}), rather than raw motion, which is then utilised by the action policy.

\subsection{Problem Setup and Architecture}
\label{sec:problem_formulation}
\label{sec:architecture}

Our problem setting is language-conditioned imitation learning, where a policy learns to predict robot actions from an expert dataset $\mathcal{D}=\{(\ell_i,\{o_t,a_t\}_{t=0}^{T_i-1})\}_{i=1}^{N}$, where $\ell_i$ is a language instruction, $o_t=(I_t^{1:C},s_t)$ denotes the RGB images from $C$ camera views and robot proprioception $s_t$, and $a_t$ is the demonstrated action. 
At each control step $t$, we aim to model a causal policy $\pi_\theta$ that predicts a distribution over an action horizon of length $H$:
\begin{equation}
    \pi_\theta(A_t \mid o_{\leq t},\ell), \qquad A_t=(a_t,\ldots,a_{t+H-1}).
\end{equation}

We formulate ELMO as a hierarchical encoder-decoder architecture that jointly trains a high-level VLA planner~$\pi_\theta^{\mathrm{plan}}$, a low-level observation-conditioned action policy~$\pi_\theta^{\mathrm{act}}$, and a dynamic routing mechanism~$\pi_\theta^{\mathrm{route}}$, forming a single end-to-end autoregressive model that uses the same causal flow during both training and inference. 

Observations first pass through a recurrent state encoder, whose output is monitored by the router $\pi_\theta^{\mathrm{route}}$ at every step $t$ to predict a binary boundary decision $b_t \in \{0,1\}$ (with $b_0 = 1$). 
Rather than querying the planner at a fixed stride, the steps where $b_t = 1$ define a sequence of chunk boundary times $\tau_k = \min\{t > \tau_{k-1} : b_t = 1\}$, partitioning the trajectory into variable-length behavioural segments. 
The inner VLA planner $\pi_\theta^{\mathrm{plan}}$ is executed \emph{only} at these selected boundary steps $t = \tau_k$, combining the encoder state with multi-view visual and language tokens to produce a new grounded semantic code $z_k$:
\begin{equation}
    z_k \sim \pi_\theta^{\mathrm{plan}}(\cdot \mid o_{\leq \tau_k}, \ell) \quad \text{at } t=\tau_k.
    \label{eq:hierarchical_formulation_planner}    
\end{equation}
At every control step $t$, let $k(t) = \max\{k : \tau_k \leq t\}$ index the most recent boundary and $z_{k(t)}$ thus denote the active grounded semantic code. 
The action policy decoder $\pi_\theta^{\mathrm{act}}$ combines $z_{k(t)}$ with the live encoder state and local visual feedback to predict an updated action window $A_t$ in a receding-horizon manner, decoupling the variable chunk duration $\tau_{k+1}-\tau_k$ from the prediction horizon $H$:
\begin{equation}
    A_t \sim \pi_\theta^{\mathrm{act}}(\cdot \mid o_{\leq t}, z_{k(t)}) \quad \forall t.
    \label{eq:hierarchical_formulation_policy}    
\end{equation}
This design establishes two temporal pathways: a fast recurrent pathway running at the control rate and a slow causal transformer operating over dynamically learned chunks.

\paragraph{Recurrent State Encoder and Dynamic Router.}
Following H-Net~\cite{hwang2025hnet}, at each step $t$, wrist-camera tokens are spatially pooled and flattened into a compact visual summary $w_t$. 
An MLP embeds $w_t$ alongside proprioception $s_t$.
To decrease reliance on precise proprioception, we add $\sigma=0.1$ noise to it.
A recurrent state encoder $E$ composed of residual GRU--MLP blocks updates the low-level hidden state:
\begin{equation}
    e_t = f_{\mathrm{emb}}(s_t,w_t), \qquad h_t^{E} = E(e_t,h_{t-1}^{E}).
\end{equation}

To decide whether the current state warrants a latent plan update, a learned router measures the cosine dissimilarity between consecutive encoder states in a projected space:
\begin{equation}
    p_t = \frac{1-\cos(W_q h_t^{E},W_k h_{t-1}^{E})}{2}.
    \label{eq:router}
\end{equation}
We set $p_0=b_0=1$ and trigger a new chunk boundary ($b_t=1$) whenever $p_t\geq0.5$ or a maximum step timeout is reached.
Retaining the encoder states at boundary steps $\{\tau_k\}$ yields a temporally compressed sequence.

\paragraph{VLA Planner.}
At each boundary $\tau_k$, the selected encoder state $h_{\tau_k}^E$ is augmented with a learned vector of additional dimensions, widening it to the planner width, to form a query, following H-Net~\cite{hwang2025hnet}. 
This query drives two cross-attention blocks that attend over the full set of current static- and wrist-camera tokens and the language instruction $\ell$, all of which are jointly encoded by the SmolVLM2~\cite{marafioti2025smolvlm} backbone. 
To preserve spatial structure during readout, visual tokens are enriched with 2D coordinate encodings. 
The resulting multimodal boundary token $u_k$ is processed by a causal transformer $T$ over preceding chunks in the episode:
\begin{equation}
    z_k = \operatorname{Proj}\left[\operatorname{RMSNorm}\bigl(T(u_0,\ldots,u_k)_k\bigr)\right],
\end{equation}
where $\operatorname{Proj}$ removes the dimensions added at the start of the planner, truncating the representation back to the encoder width. 
The vector $z_k$ serves as a continuous grounded semantic code that summarizes behavioural intent and long-horizon task progress at a compressed temporal resolution.

\paragraph{Action Policy.}
The observation-conditioned action policy translates the active grounded semantic code $z_{k(t)}$ into raw actions. 
At each step $t$, the active $z_{k(t)}$ is combined with the live encoder state $h_t^E$ via a residual skip connection~\cite{hwang2025hnet} to form the decoder input $d_t$.
A recurrent decoder processes $d_t$, and its output cross-attends to the wrist-camera to supply fine-grained, up-to-date local visual feedback for reactive manipulation while the high-level code remains fixed.
The resulting feature $h_t^D$ conditions a flow-matching action head $v_\theta(X_t^u, u, h_t^D)$ to generate the action window $A_t$. 
Specifically, for flow time $u \in [0,1]$ and Gaussian noise $\epsilon_t \sim \mathcal{N}(0,I)$, the action head predicts the velocity field transporting $\epsilon_t$ to $A_t$ along the linear interpolation path $X_t^u = (1-u)\epsilon_t + uA_t$; during inference, $A_t$ is sampled by integrating $v_\theta$ from $u=0$ to $u=1$ over five Euler steps.

\subsection{Training and Inference}
\label{sec:training}

We jointly optimize the recurrent state encoder, router, VLA planner, action policy decoder, and action head end-to-end while keeping the visual backbone frozen.

\paragraph{Boundary-Weighted Flow Matching.}
During training, we sample $u\sim\mathcal{U}[0,1]$ and $\epsilon_t \sim \mathcal{N}(0,I)$ to train the action head $v_\theta(X_t^u,u,h_t^D)$ to predict the target velocity $A_t-\epsilon_t$.
Let $l_{t,j}$ denote the mean squared velocity error at horizon offset $j$.
We minimize the boundary-weighted flow-matching objective:
\begin{equation}
    \mathcal{L}_{\mathrm{act}} = \mathbb{E}_{u,\epsilon}\left[\frac{\sum_{t,j}m_{t,j}w_t l_{t,j}}{\sum_{t,j}m_{t,j}w_t}\right], \,\ w_t=1+\alpha\,\operatorname{sg}(b_t)
    \label{eq:action_loss}
\end{equation}
where $m_{t,j}$ masks invalid steps, $\operatorname{sg}(\cdot)$ is the stop-gradient operator, and $\alpha=4$ upweights action windows predicted immediately at chunk boundaries so the VLA planner learns informative updates.
To allow action gradients to train the discrete router, a confidence-weighted straight-through estimator propagates gradients to $p_t$.

\paragraph{Router Regularisation.}
To prevent trivial routing, we use the H-Net approach~\cite{hwang2025hnet} and regularise the router toward a target compression ratio of $C_\text{ratio}$ control steps per chunk:
\begin{equation}
    \mathcal{L}_{\mathrm{ratio}} = \frac{C_\text{ratio}}{C_\text{ratio}-1}\left[(C_\text{ratio}-1)FG+(1-F)(1-G)\right],
\end{equation}
where $F$ is the empirical fraction of selected boundaries and $G$ is the mean boundary probability across the batch. 
During early training, to improve routing stability, we also guide the router with a binary cross-entropy warm-start loss $\mathcal{L}_{\mathrm{warm}}$ derived from heuristic kinematic boundaries, whose weight, $w_{\mathrm{warm}}$, linearly decays over the first $S_{\mathrm{warm}}$ steps:
\begin{equation}
    \mathcal{L} = \mathcal{L}_{\mathrm{act}}+\lambda_{\mathrm{ratio}}\mathcal{L}_{\mathrm{ratio}}+w_{\mathrm{warm}}\mathcal{L}_{\mathrm{warm}}.
\end{equation}

\paragraph{History Regularization.}
We incorporate two data-level mechanisms during training to address the drawbacks of history-conditioned policies.
First, unlike language, robot behaviour is highly autocorrelated, which makes it easy to achieve low loss simply by extrapolating from recent history~\cite{pertsch2025fast}. 
This is especially problematic for the VLA planner, as much of the plan can be inferred from history once the episode has started.
We therefore do \textit{onset-prioritised sampling} and focus training on episodes where the planner struggles to choose the correct initial plan, forcing it to rely more heavily on visual and task information.  
We track an exponential moving average $L_i$ of each demonstration's unweighted loss over its first $K_{\mathrm{init}}$ steps and sample training trajectories with probability
\begin{equation}
    q_i = \frac{1}{2N}+\frac{1}{2}\frac{L_i^2}{\sum_{r=1}^{N}L_r^2},
    \label{eq:loss_sampling}
\end{equation}
balancing uniform dataset coverage with focused updates on difficult task openings. 

Second, the policy can use its history to store object locations from earlier in the rollout, removing the need to re-observe the scene.
While this shortcut succeeds during training, it makes the policy brittle during inference.
We therefore break single-trajectory continuity by constructing training sequences from spliced fragments of different demonstrations that share the same instruction $\ell$, without resetting hidden states or chunk history at the joins. 
Target masks $m_{t,j}$ prevent supervision windows from crossing artificial splice points.
This limits the policy from relying on memory for long-term state tracking while still allowing it to use short-term history for smooth, precise execution.

\paragraph{Inference.}
At each step $t$, the encoder updates $h_t^E$ and the router evaluates Eq.~\eqref{eq:router}.
The VLA planner predicts a new latent plan only when the router places a new boundary. 
The flow head integrates noise over five Euler steps to produce $A_t$, and the robot executes the first action $a_t$ before acquiring $o_{t+1}$.

\section{Experiments}
\label{sec:experiments}

We evaluate ELMO across two standard manipulation benchmarks (LIBERO~\cite{liu2023libero} and CALVIN~\cite{mees2021calvin}), a zero-shot robustness benchmark (LIBERO-Plus~\cite{fei2025liberoplus}), a new shortcut-free compositional benchmark (Blocks), and a physical bimanual robot. 
We first analyze the learned latent spaces~(\Cref{sec:latent_analysis}), showing that while prior action codecs primarily reparameterize low-level kinematics, ELMO learns grounded semantic codes that capture task intent, target identity, and execution context. 
We then demonstrate that this structured abstraction allows us to outperform alternative latent codes paired with a similar-sized backbone across standard and perturbed simulation benchmarks~(\Cref{sec:simulation-results}), enables instruction following where existing methods fail~(\Cref{sec:blocks-v2}), and transfers effectively to real hardware~(\Cref{sec:real_results}).

\subsection{Benchmarks and Baselines}
\label{sec:datasets}

\paragraph{Benchmarks.}
We first evaluate on two established simulation benchmarks: LIBERO~\cite{liu2023libero} and CALVIN~\cite{mees2021calvin} (see supplementary for details). 
LIBERO comprises four suites---Spatial, Object, Goal, and LIBERO-10---targeting distinct facets of language-conditioned generalization, while CALVIN ($\text{ABCD}\rightarrow\text{D}$) evaluates long-horizon execution and chaining of independent skills. 
However, studies show that models can solve these standard benchmarks via spurious shortcuts—inferring task intent from initial scene layouts or gripper proximity rather than genuinely grounding the language instruction---leaving them brittle under distribution shifts~\cite{jiang2026we, pmlr-v305-xing25a, fei2025liberoplus, sendai2026minervasmallmanipulationpolicy}. 
To rigorously test robustness and causal instruction following, we additionally evaluate zero-shot on LIBERO-Plus~\cite{fei2025liberoplus}, which applies perturbations to LIBERO, and introduce our shortcut-free Blocks benchmark~(\Cref{sec:blocks-v2}).

\paragraph{Baselines.}
We compare ELMO against SmolVLA~\cite{shukor2025smolvla}, which directly predicts continuous actions, and \mbox{MINT-30M}~\cite{huang2026mimic}, a similarly sized hierarchical policy using latent actions. 
We also compare against alternative latent action tokenizers combined with SmolVLA~\cite{shukor2025smolvla}.
Following the protocol of EDAR~\cite{xu2026edar}, we evaluate pretrained action codecs---FAST~\cite{pertsch2025fast} and EDAR~\cite{xu2026edar}---conditioned on the same SmolVLM2~\cite{marafioti2025smolvlm} backbone (see supplementary for implementation details). 
All methods are evaluated closed-loop on identical test seeds across $3$ training seeds.

\subsection{Latent Space Analysis}
\label{sec:latent_analysis}

\begin{figure}[tbp]
    \centering
    \includegraphics[trim={0cm 0cm 15cm 0.4cm},clip,width=1.0\linewidth]{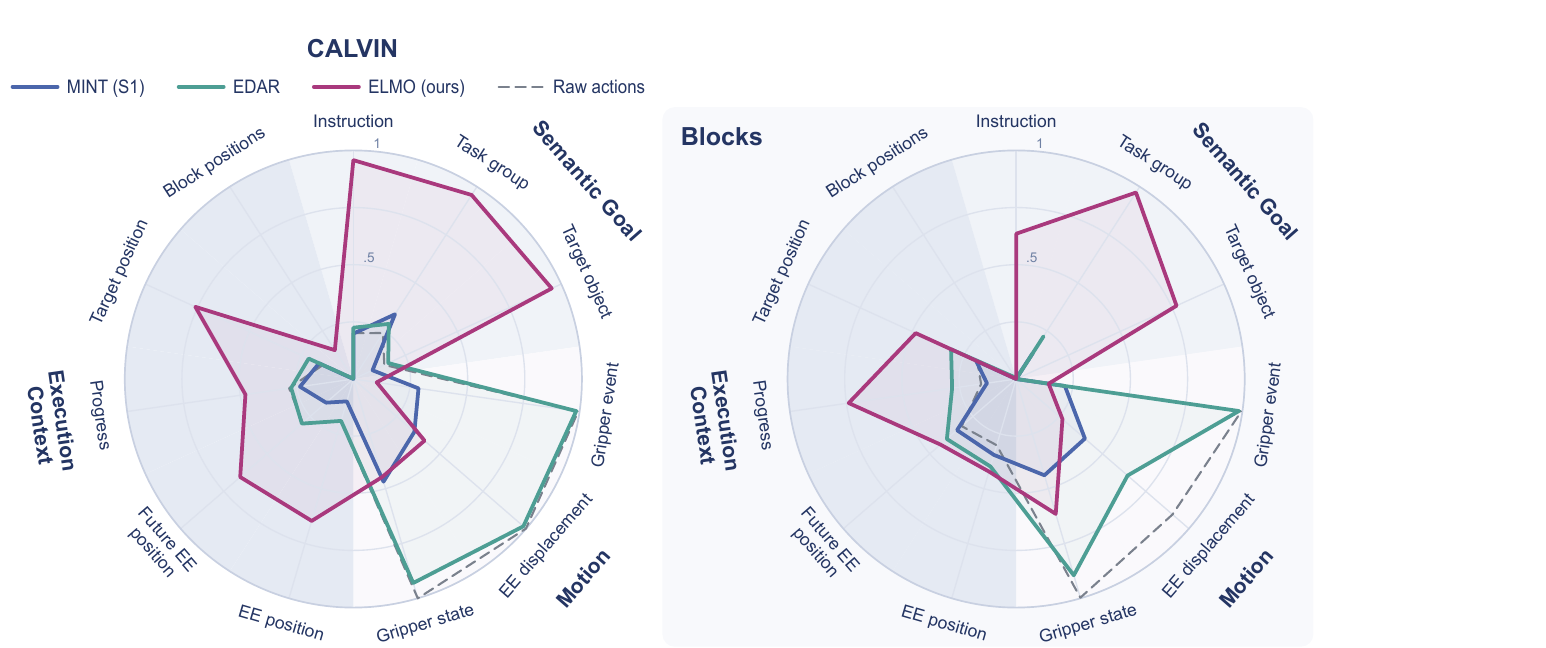}
    \caption{Linear probe performance across kinematic and semantic attributes. ELMO encodes more semantic and context information and less motion than all baselines.}
    \label{fig:radar_charts}
\end{figure}

\begin{figure}[tbp]
    \centering
    \includegraphics[trim={1cm 0cm 0cm 0.85cm},clip,width=\linewidth]{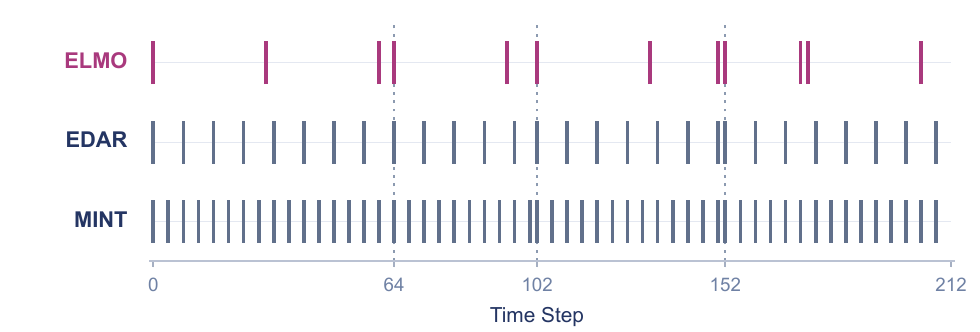}
    \caption{Dynamic chunk boundaries placed by ELMO compared to the static replan rate of MINT and EDAR along a CALVIN rollout (gray lines indicate new subtask).}
    \label{fig:chunk_boundaries_chart}
\end{figure}

\begin{figure*}[t]
    \centering
    \includegraphics[trim={0cm 0.5cm 0cm 0cm},clip,width=\linewidth]{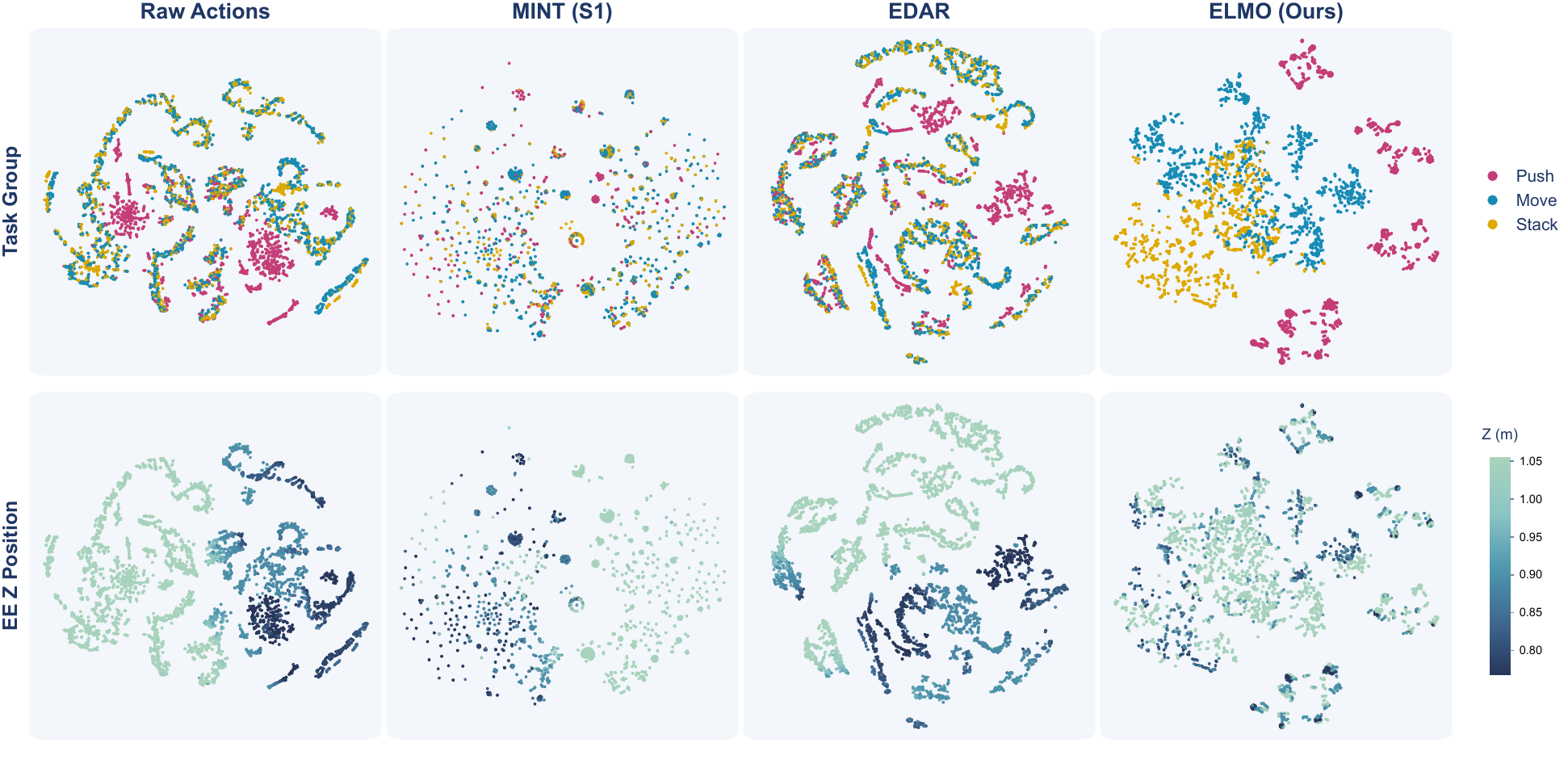}
    \caption{t-SNE projections comparing latent representations of MINT~\cite{huang2026mimic}, EDAR~\cite{xu2026edar}, and ELMO on Blocks. For MINT, we plot the single intent tokens. EDAR's t-SNE structure is very close to that of raw actions and MINT's structure shows no organization by task. ELMO shows clear organization by task. Representations overlap as both push and stack trajectories share similar approach subsequences.}
    \label{fig:tsne_charts}
\end{figure*}

To understand what information ELMO, MINT, and EDAR capture in their latent representations across LIBERO, CALVIN, and Blocks, we fit ridge-regression linear probes to latent plans extracted from demonstration trajectories (for details see supplementary). 
As shown in~\Cref{fig:radar_charts}, MINT and EDAR latents largely encode the same information as raw action---concentrated in end-effector displacements and gripper states---functioning as \emph{motion codes} that reparameterize kinematics. 
In contrast, ELMO's latents are highly predictive of target objects, active skills, and task progress, while retaining contextual execution cues beyond the static task label. 
We therefore characterize ELMO's emergent representations as \emph{grounded semantic codes}: they specify not only \emph{what} semantic skill to execute, but \emph{how} to ground it in the current scene. 
This distinction is echoed in the t-SNE plots~(\Cref{fig:tsne_charts}). Whereas MINT and EDAR latents organize by movement direction and gripper state, ELMO's latents naturally cluster by skills and target tasks without explicit contrastive supervision. 

Finally, visualizing ELMO's learned chunk boundaries~(\Cref{fig:teaser,fig:chunk_boundaries_chart}) shows that plan updates naturally align with semantic sub-phases (e.g., reaching vs.\ grasping). 
ELMO dynamically allocates shorter chunks during contact-rich, high-precision phases like grasping and longer chunks during coarse transit, utilizing high-level queries more efficiently than fixed-stride methods. 

Together, these properties yield substantial parameter efficiency without sacrificing performance. Because its latent space is both semantically grounded and adaptively segmented, ELMO outperforms baselines on manipulation benchmarks~(\Cref{sec:simulation-results}) with only 22M unfrozen parameters across the encoder, router, VLA action head, and decoder—compared to 100M in SmolVLA's action head. 

\subsection{Main Simulation Results}
\label{sec:simulation-results}

Table~\ref{tab:simulation} summarizes performance across LIBERO, LIBERO-Plus, and CALVIN. 
On LIBERO, ELMO achieves a $90.2\%$ average success rate, outperforming SmolVLA ($85.7\%$) and the best action codec (EDAR, $88.6\%$) while matching MINT-30M ($90.2\%$) without relying on image augmentations (without which MINT-30M drops to $84.4\%$). 
The advantage of grounded semantic abstraction is most pronounced on long-horizon CALVIN ($\text{ABCD}\rightarrow\text{D}$) chains, where ELMO completes an average of $4.08$ consecutive subtasks---surpassing the strongest baseline (SmolVLA + FAST) by $+0.75$ subtasks.

On LIBERO-Plus, ELMO achieves $64.0\%$ zero-shot success.
While MINT-30M achieves $61.0\%$, it requires using image augmentation to do so.
When comparing across methods without such augmentations, ELMO outperforms the next best method by $9.6$ percentage points, with the largest gains under Camera, Language, and Noise perturbations (see supplementary for details).

\subsection{Compositional Grounding and Generalization}
\label{sec:blocks-v2}

To isolate instruction grounding from scene-layout shortcuts in a real-world-transferable setting~(see Table in supplementary), we introduce Blocks, a bimanual ALOHA MuJoCo benchmark~\cite{todorov2012mujoco, zhao2024aloha} featuring two opposing ViperX arms, three RGB cameras, and five distinctly coloured cubes placed randomly in every scene. 
Tasks span three families---directional pushing, stacking, and regional placement---with targets sampled independently of gripper proximity, occasionally requiring cross-table bimanual handovers. 
We generate $10{,}000$
clean scripted demonstrations and augment them with $2{,}500$ recovery trajectories in which an initially misaligned approach is actively corrected.

Together, these design choices force policies to rely on genuine language grounding rather than spatial shortcuts, providing a rigorous measure of generalization. 
Compounding this challenge, the $5\text{ cm}$ cubes afford only a $0.7\text{ cm}$ total tolerance within the $5.7\text{ cm}$ gripper opening, requiring fine-grained manipulation precision.

As shown in Table~\ref{tab:blocks_results}, all baseline policies collapse on Blocks ($\le 5.0\%$ ID, $\le 2.0\%$ OOD), failing to disambiguate which of the five cubes to manipulate when proximity shortcuts are removed. 
In contrast, ELMO achieves $38.6\%$ ID and $31.7\%$ OOD success by grounding the target colour in its high-level plan and purposefully approaching the instructed block.
Remaining failures occur during fine-grained motion rather than target selection, reflecting the challenge of learning high-precision contact.

\begin{table*}[t]
  \centering
  \caption{\textbf{Main simulation benchmark results.} LIBERO and LIBERO-Plus report average episode success rates (\%). CALVIN reports average consecutively completed subtasks (out of 5). Entries report mean $\pm$ standard error across $3$ seeds.}
  \label{tab:simulation}
  \resizebox{\textwidth}{!}{%
  \begin{tabular}{lccccccc}
    \toprule
    & LIBERO $\uparrow$
    & \multicolumn{5}{c}{LIBERO-Plus $\uparrow$}
    & CALVIN ABCD$\rightarrow$D $\uparrow$ \\
    \cmidrule(lr){3-7}
    Method & Avg. & Spatial & Object & Goal & Long & Avg. & Avg. Len. \\
    \midrule
    SmolVLA~\cite{shukor2025smolvla} & $85.7 \pm 0.6$ & $56.4 \pm 0.9$ & $56.1 \pm 0.7$ & $51.7 \pm 1.1$ & $36.9 \pm 1.3$ & $50.2 \pm 0.8$ & $3.07 \pm 0.03$ \\
    SmolVLA + FAST codec~\cite{pertsch2025fast} & $83.4 \pm 1.2$ & $55.5 \pm 1.1$ & $59.4 \pm 1.7$ & $48.1 \pm 0.6$ & $37.9 \pm 1.6$ & $50.1 \pm 0.3$ & $3.33 \pm 0.05$ \\
    SmolVLA + EDAR codec~\cite{xu2026edar} & $88.6 \pm 0.8$ & $63.0 \pm 0.5$ & $61.5 \pm 0.7$ & $52.8 \pm 0.8$ & $40.7 \pm 0.9$ & $54.4 \pm 0.2$ & $3.32 \pm 0.02$ \\
    \midrule
    MINT-30M~\cite{huang2026mimic} & $\mathbf{90.2 \pm 0.8}$ & $65.9 \pm 0.9$ & $70.8 \pm 1.1$ & $\mathbf{61.1 \pm 0.9}$ & $46.5 \pm 2.4$ & $61.0 \pm 1.3$ & $2.68 \pm 0.11$ \\
    MINT-30M (w/o image aug.)~\cite{huang2026mimic} & $84.4 \pm 0.7$ & $54.2 \pm 1.9$ & $53.0 \pm 2.4$ & $50.9 \pm 1.4$ & $33.4 \pm 0.9$ & $47.8 \pm 1.3$ & -- \\
    \midrule
    \textbf{ELMO (Ours)} & $\mathbf{90.2 \pm 0.4}$ & $\mathbf{72.6 \pm 1.0}$ & $\mathbf{71.9 \pm 1.1}$ & $58.9 \pm 0.1$ & $\mathbf{53.2 \pm 1.5}$ & $\mathbf{64.0 \pm 0.8}$ & $\mathbf{4.08 \pm 0.04}$ \\
    \bottomrule
  \end{tabular}}

  \vspace{0.6em}

  \begin{minipage}[c]{0.48\textwidth}
      \centering
      \captionof{table}{\textbf{Blocks benchmark results.} Success rate (\%) on in-distribution (ID) vs.\ held-out colour--position grid cells (OOD). Entries report mean $\pm$ std error across 3 seeds.}
      \label{tab:blocks_results}
      \small
      \begin{tabular}{lcc}
          \toprule
          Method & ID (\%) $\uparrow$ & OOD (\%) $\uparrow$ \\
          \midrule
          SmolVLA~\cite{shukor2025smolvla} & $4.9 \pm 0.7$ & $0.7 \pm 0.4$ \\
          SmolVLA + FAST codec~\cite{pertsch2025fast} & $3.2 \pm 0.4$ & $0.5 \pm 0.1$ \\
          SmolVLA + EDAR codec~\cite{xu2026edar} & $4.7 \pm 0.4$ & $2.0 \pm 0.5$ \\
          MINT-30M~\cite{huang2026mimic} & $2.0 \pm 0.6$ & $1.1 \pm 0.3$ \\
          \midrule
          \textbf{ELMO (Ours)} & $\mathbf{38.6 \pm 2.9}$ & $\mathbf{31.7 \pm 5.1}$ \\
          \bottomrule
      \end{tabular}
  \end{minipage}
  \hfill
  \begin{minipage}[c]{0.48\textwidth}
      \centering
      \includegraphics[trim={5cm 7cm 5cm 7cm},clip,width=0.75\linewidth]{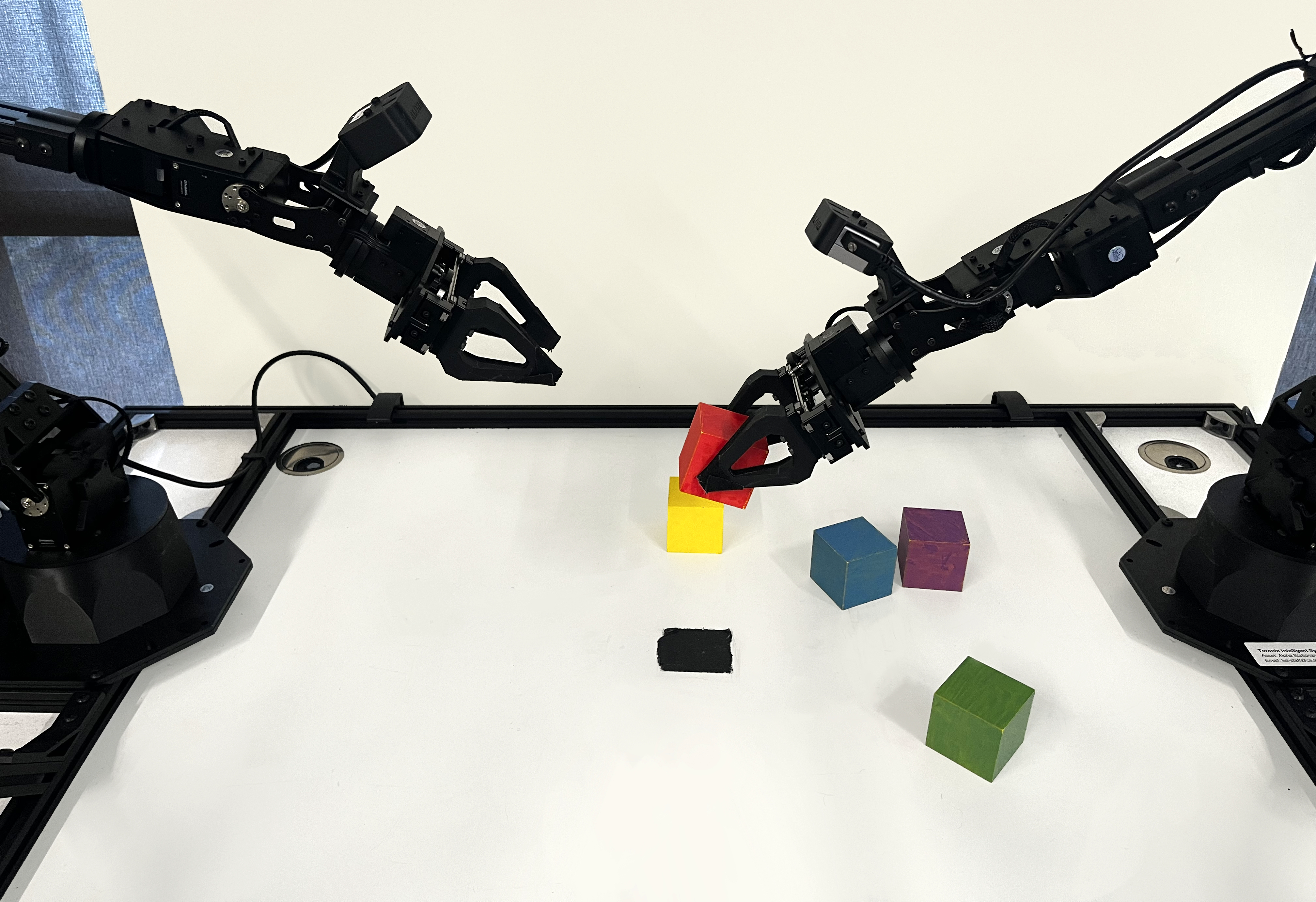}
      \captionof{figure}{Real-world bimanual ALOHA evaluation setup~\cite{zhao2024aloha}.}
      \label{fig:real_setup}
  \end{minipage}
\end{table*}

\subsection{Real-World Evaluation}
\label{sec:real_results}

We deploy ELMO on a physical bimanual ALOHA setup mirroring our Blocks benchmark~(\Cref{fig:real_setup}). 
Given the low success rates observed for stacking in simulation, we focus our real-world evaluation on the pushing and regional placement families. 
The two task families, five coloured blocks, and four directions yield $40$ distinct task combinations. 
We fine-tune each policy from its simulation checkpoint for $10\text{k}$ steps using $30$ human demonstrations per combination, and evaluate across $25$ randomized pushing and placement trials ($50$ rollouts total; see supplementary for full training and deployment details). 
Overcoming substantial sim-to-real discrepancies in camera viewpoints, scene appearance, and physical dynamics, ELMO achieves $13/25$ on push trials and $8/25$ on regional placement trials, outperforming SmolVLA ($2/25$ push, $0/25$ placement).

\subsection{Ablation Studies}
\label{sec:ablations}
Ablations train on CALVIN with 16,000 steps and 3 seeds. 

\paragraph{Router Warm-Start.}
Removing the kinematic router warm-start ($\mathcal{L}_{\mathrm{warm}}$) lowers average completed tasks only slightly, from 4.04 to 3.90, but it changes where the router places its boundaries.
On held-out demonstrations, the warm-started router cuts within 2 steps of 82\% of grasps, $3.8\times$ the 21\% expected from the same number of randomly placed boundaries, and the nearest cut usually falls exactly on the grasp.
Without the warm-start this drops to 39\%. 
The remaining cuts tend to come late, up to 11 steps after the grasp.
The router thus discovers some event-aligned boundaries on its own, but the warm-start makes boundary placement at contact events consistent and precise.

\paragraph{Low-Level Feedback and Data Augmentation.}
Removing control-rate wrist cross-attention or the encoder skip connection reduces the average number of completed tasks by 1.05 and 0.58 respectively, demonstrating that continuous low-level visual feedback is critical for the decoder.
Disabling trajectory splicing and onset-prioritized sampling reduces average completed tasks by 0.79 as the model over relies on its history and ignores the visual observations.

\paragraph{Compression and Dynamic Routing.}
Replacing the learned router with fixed-length chunks while maintaining $C_\text{ratio}=12$ reduces average completed tasks from 4.04 to 3.89.
This means dynamic routing does help align the chunks with current intentions but is not critical.

\section{Conclusion and Limitations}
\label{sec:conclusion}

We present \textbf{ELMO}, an end-to-end hierarchical VLA policy that avoids the collapse of prior latent action model representations into \emph{motion codes}. 
By pairing dynamic temporal routing with an observation-conditioned flow-matching decoder, ELMO learns \emph{grounded semantic codes} that capture task-level behavioural intent and environmental context while delegating reactive control to a lightweight action policy. 
On both LIBERO-Plus and CALVIN, ELMO achieves the best performance among the evaluated models with a fraction of the trainable parameters, and uniquely succeeds at instruction following on our shortcut-free Blocks benchmark in both simulation and the real world.

\paragraph{Limitations.}
First, while dynamic routing emerges end-to-end, stable boundary alignment with contact events still benefits from an early kinematic warm-start ($\mathcal{L}_{\mathrm{warm}}$) and a predefined target compression ratio $C_\text{ratio}$. 
Second, on Blocks, although ELMO resolves target disambiguation, it still struggles with sub-centimeter grasping and stacking precision—suggesting the need for closed-loop RL fine-tuning or higher-resolution local sensing. 
Finally, extending ELMO's recurrent state encoder to cross-embodiment pretraining and unlabeled video data remains an important direction for future work.

\clearpage
\section{LLM Usage Statement}
\label{sec:llm_usage_statement}
LLM agents were used in the research process to write code, create visualizations, and monitor experiments. 
All results were manually verified by the authors. 
LLMs were also used to refine the writing in this paper.

\section{Acknowledgements}
\label{sec:acknowledgements}
This research was supported by the \href{https://vectorinstitute.ai/}{Vector Institute} and  \href{https://alliancecan.ca/}{Digital Research Alliance of Canada}.
We would also like to thank Shuhong Zheng for his warm support.

{\small
\bibliographystyle{ieeenat_fullname}
\bibliography{main}
}
\clearpage
\setcounter{page}{1}
\maketitlesupplementary

\setcounter{section}{0}
\renewcommand{\thesection}{\Alph{section}}

\section{ELMO Implementation \& Training Details}
\label{sec:supp_implementation}

In CALVIN and LIBERO, $o_t$ pairs one scene and one wrist camera with an 8D state $s_t$, and $a_t$ is 7D (relative end-effector pose and binary gripper); in Blocks, $o_t$ pairs a cropped overhead and two wrist cameras with a 20D state $s_t$, and $a_t$ is 14D (bimanual joint-position deltas and absolute gripper targets). Both $s_t$ and $a_t$ are z-score normalized.
Images are encoded by the frozen SmolVLM2 vision tower into an $8\times8$ token grid per camera. In the control-rate encoder $E$, wrist tokens are pooled to $4\times4$, projected to 32D, flattened into $w_t$, and embedded with $s_t$ via a 2-layer MLP $f_{\mathrm{emb}}$.
At each boundary $\tau_k$, the frozen first 16 SmolVLM2 language layers process the boundary camera tokens (with start/end delimiters) and instruction $\ell$, projecting the 960D output to $D_{\mathrm{main}}=384$ and adding linear camera-ID and sinusoidal 2D grid encodings. The extended boundary state $h_{\tau_k}^E$ queries these tokens via pre-norm cross-attention; the resulting $u_k$ adds a learned chunk-position embedding and passes through the causal transformer $T$.
The decoder input is $d_t = \bar{z}_{k(t)} + W_{\mathrm{skip}}h_t^E$ ($W_{\mathrm{skip}}$ zero-initialized, $W_q, W_k$ identity-initialized). Decoder wrist cross-attention keys add learned positional embeddings and zero-initialized camera-ID and 2D coordinate projections. All remaining architectural and optimization hyperparameters are listed in~\Cref{tab:hyperparams}.
Ablations on CALVIN and Blocks were trained for 16,000 steps with otherwise default parameters.

\paragraph{Heuristic Router Warm-Start.}
For $\mathcal{L}_{\mathrm{warm}}$ ($w_{\mathrm{warm}}$ decaying linearly from $1$ to $0$ over $S_{\mathrm{warm}}$), target boundaries are extracted causally from raw proprioception $s_t$. A step $t$ is a candidate boundary if any gripper state crosses its open/close threshold between $t-1$ and $t$, or if the arm speed $v_t = \lVert s_t^{\mathrm{arm}} - s_{t-1}^{\mathrm{arm}} \rVert$ (computed over end-effector position on CALVIN/LIBERO and arm joint positions on Blocks) reaches a local minimum at $t-1$. A local minimum requires $v_{t-1} < v_{t-2}$, $v_{t-1} \le v_t$, and $v_{t-1}$ to fall below the 25th percentile of training per-step speeds (computed over moving steps only on Blocks). Candidates within 4 steps of a prior heuristic boundary are discarded.

\begin{table}[h]
  \centering
  \caption{\textbf{ELMO architecture and training hyperparameters.}}
  \label{tab:hyperparams}
  \resizebox{\columnwidth}{!}{%
  \begin{tabular}{ll}
    \toprule
    \textbf{Hyperparameter} & \textbf{Value} \\
    \midrule
    Action horizon $H$ & $16$ \\
    Encoder / decoder width $D_{\mathrm{enc}}$ & $256$ (pre-norm residual GRU--MLP, RMSNorm) \\
    Planner width $D_{\mathrm{main}}$ & $384$ \\
    Boundary cross-attention & $6$ heads \\
    Planner causal transformer $T$ & $4$ blocks, width $384$, $6$ heads \\
    Decoder wrist cross-attention & $4$ heads ($8\times8$ grid), RMSNorm \\
    Flow-matching head $v_\theta$ & $3$-layer MLP, width $1024$ \\
    \midrule
    Target compression ratio $C_{\text{ratio}}$ & $12$ (max timeout: $30$ steps) \\
    Ratio loss weight $\lambda_{\mathrm{ratio}}$ & $0.03$ \\
    Router warm-start $S_{\mathrm{warm}}$ & $5{,}000$ updates ($w_{\mathrm{warm}}: 1 \to 0$) \\
    Onset window $K_{\mathrm{init}}$ / EMA decay & $16$ steps / $0.7$ (unseen init.\ to mean) \\
    Random sequence start prob. & $0.5$ \\
    Spliced fragments per sequence & $2$--$3$ \\
    Max spliced length / min fragment & $192$ / $8$ (CALVIN); $256$ / $64$ (LIBERO, Blocks) \\
    \midrule
    Optimizer & AdamW ($\beta=(0.9,0.95)$, WD $0.1$, grad clip $1.0$) \\
    Batch size & $32$ \\
    Base learning rate & $2\times10^{-4} \to 2\times10^{-5}$ (cosine w/ restarts) \\
    Control-rate LR multiplier & $\sqrt{C_{\text{ratio}} D_{\mathrm{main}}/D_{\mathrm{enc}}} \approx 4.24$ \\
    Cycle length & $8{,}000$ updates \\
    Cycle warm-up & $1{,}000$ steps (cycle 1); AdamW state reset + $500$ steps \\
    Total training updates & $16{,}000$ (LIBERO, CALVIN); $40{,}000$ (Blocks) \\
    \bottomrule
  \end{tabular}}
\end{table}

\section{Benchmark Protocols}
\label{sec:supp_benchmarks}

We evaluate on LIBERO~\cite{liu2023libero}, LIBERO-Plus~\cite{fei2025liberoplus}, CALVIN~\cite{mees2021calvin}, and our Blocks benchmark.
We describe the training data, policy interfaces, evaluation procedures, and baseline implementations below. 
Unless otherwise specified, variation across seeds refers to independently trained policies.
For baselines, which have a separate action tokenizer training step, representation models are shared across policy seeds.

\subsection{LIBERO}
\label{sec:supp_libero}

\paragraph{Training data.}
We train jointly on the four standard suites: LIBERO-Spatial, LIBERO-Object, LIBERO-Goal, and LIBERO-10, also referred to as
LIBERO-Long. Each suite contains ten tasks.
The baseline dataset is \texttt{lerobot/libero}~\cite{cadene2024lerobot}, containing 1,693 demonstrations across the four suites.
We do not use LIBERO-90 demonstrations for any training run on LIBERO.

\paragraph{Environment initialization and termination.}
Both cameras render at $256\times256$ resolution.
Images are rotated by $180^\circ$ to match the training and LeRobot
convention.
Each episode resets to its specified benchmark initial state, followed by ten settling steps with zero arm motion and an open-gripper command.
The policy state and action queue are reset between episodes.
An episode succeeds when the simulator's task predicate becomes true within the task's step budget.

\begin{table}[t]
    \centering
    \small
    \begin{tabular}{lc}
        \toprule
        Suite & Maximum policy steps \\
        \midrule
        LIBERO-Spatial & 280 \\
        LIBERO-Object & 280 \\
        LIBERO-Goal & 300 \\
        LIBERO-10 / Long & 520 \\
        \bottomrule
    \end{tabular}
    \caption{Evaluation step budgets for LIBERO.}
    \label{tab:supp_libero_limits}
\end{table}

Evaluation uses the default simulator control frequency of $20$\,Hz.

\paragraph{Evaluation sample counts.}
We evaluate indices $0$--$24$ for each model, totaling $4\times10\times25=1{,}000$ episodes per seed.

We report success percentages for each suite and their mean.
Uncertainty across policies is computed across training seeds.

\subsection{LIBERO-Plus}
\label{sec:supp_libero_plus}

\paragraph{Zero-shot robustness evaluation.}
We evaluate the clean-LIBERO policies on LIBERO-Plus~\cite{fei2025liberoplus} without additional training.
The full benchmark contains 10,030 perturbed task variants.
We execute one trial per variant using initial-state index zero, with $256\times256$ camera rendering.
Each method retains its clean-LIBERO action-execution horizon and uses the corresponding base suite's step budget.

\begin{table}[t]
    \centering
    \small
    \begin{tabular}{lr}
        \toprule
        Perturbation category & Variants \\
        \midrule
        Camera viewpoints & 1,599 \\
        Robot initial states & 1,550 \\
        Language instructions & 1,537 \\
        Lighting & 1,142 \\
        Background textures & 1,076 \\
        Sensor noise & 1,601 \\
        Object layouts & 1,525 \\
        \midrule
        Total & 10,030 \\
        \bottomrule
    \end{tabular}
    \caption{Composition of the full LIBERO-Plus evaluation.}
    \label{tab:supp_plus_categories}
\end{table}

\paragraph{Instruction handling.}
For non-language perturbations, the upstream task-language field can include metadata derived from the perturbation filename.
For example, an instruction may contain camera or initialization identifiers following the actual task description.
We remove this metadata and recover the canonical training instruction.
For language-perturbation variants, we preserve the intended rewritten instruction.

\paragraph{Aggregation.}
We report success within each perturbation category and overall success pooled across task variants.
The overall score is task-weighted:
\begin{equation}
    \mathrm{SR}_{\mathrm{Plus}}
    =
    \frac{1}{10{,}030}
    \sum_{i=1}^{10{,}030}
    \mathbb{I}[\text{variant }i\text{ succeeds}].
\end{equation}
It is not the unweighted mean of the seven category percentages.
For multiple training seeds, we report the mean and standard error across seeds.

\subsection{CALVIN}
\label{sec:supp_calvin}

\paragraph{Training split.}
We evaluate on the CALVIN~\cite{mees2021calvin} ABCD$\rightarrow$D split.

We use \texttt{fywang/calvin-task-ABCD-D-lerobot}~\cite{cadene2024lerobot}.
The combined dataset contains 24,053 episodes.
Training is restricted to episodes $0$--$22{,}965$, comprising 22,966 episodes and 1,357,600 frames.
Episodes $22{,}966$--$24{,}052$ belong to the validation split and are excluded from training.

\paragraph{Observations and actions.}
The dataset provides a $200\times200$ static-camera image, an $84\times84$ gripper-camera image, a 15-dimensional robot observation, and a seven-dimensional relative action.
The baseline policies use the dataset's 15-dimensional robot state.

ELMO transforms this state to eight dimensions: three TCP position values, three axis-angle orientation values, gripper opening width, and gripper command.
Orientation is expressed relative to a fixed reference rotation, avoiding the discontinuity of an absolute axis-angle representation near $\pi$.
Both baselines and ELMO output CALVIN's seven-dimensional relative action, with a binary gripper command.

\paragraph{Chained-task evaluation.}
Each evaluation chain contains five subtasks.
The environment is initialized once to the chain's specified initial
condition.
For each subtask, the policy receives its validation instruction and has a budget of 360 control steps.
Policy memory and action queues are reset at each subtask transition, while the physical environment continues from the preceding successful subtask.
Evaluation stops at the first failed subtask or after all five subtasks succeed.

Let $L_i\in\{0,\ldots,5\}$ denote the number of consecutively completed subtasks in chain $i$.
We report
\begin{align}
    \mathrm{SR}_k
    &=
    \frac{1}{N}
    \sum_{i=1}^{N}\mathbb{I}[L_i\geq k],
    \qquad k\in\{1,\ldots,5\},\\
    \mathrm{AvgLen}
    &=
    \frac{1}{N}\sum_{i=1}^{N}L_i
    =
    \sum_{k=1}^{5}\mathrm{SR}_k.
\end{align}

\paragraph{Chain selection.}
The evaluator uses the official 1,000-chain list to select the first 500 entries for evaluation.

\subsection{Shared SmolVLA Backbone}
\label{sec:supp_shared_backbone}

SmolVLA~\cite{shukor2025smolvla}, FAST~\cite{pertsch2025fast}, and EDAR~\cite{xu2026edar} use the LeRobot 0.6.1 SmolVLA implementation.
They load the pretrained \texttt{SmolVLM2-500M-Video-Instruct}~\cite{marafioti2025smolvlm} vision-language backbone and initialize a fresh action expert.
The vision-language backbone is frozen.
The action expert, state projection, and representation-specific output components are trained.

The backbone uses the first 16 VLM layers.
This initialization supplies pretrained vision-language weights rather than a robot-action-pretrained SmolVLA policy.
State and action preprocessing uses mean/std normalization unless a representation performs an additional transformation internally.
These three baseline families train without image augmentation.

\begin{table*}[t]
    \centering
    \small
    \begin{tabular}{lccc}
        \toprule
        Setting & SmolVLA & FAST & EDAR \\
        \midrule
        Policy batch size & 64 & 64 & 64 \\
        Output representation
            & Raw action chunk
            & Autoregressive discrete tokens
            & Four continuous registers \\
        Predicted action horizon, LIBERO/CALVIN & 50 & 10 & 16 \\
        Executed actions per query & 10 & 10 & 8 \\
        Peak policy learning rate
            & $10^{-4}$ & $1.5\times10^{-4}$ & $1.5\times10^{-4}$ \\
        Final schedule learning rate
            & $2.5\times10^{-6}$ & $5\times10^{-6}$ & $5\times10^{-6}$ \\
        Schedule length & 100k & 100k & 100k \\
        Warmup updates & 1k & 1k & 1k \\
        \bottomrule
    \end{tabular}
    \caption{Principal SmolVLA-based policy settings on LIBERO and
    CALVIN. Values were taken from papers wherever possible.}
    \label{tab:supp_baseline_settings}
\end{table*}

\subsection{SmolVLA}
\label{sec:supp_smolvla}

SmolVLA directly predicts a 50-step action chunk using its flow-matching action expert.
At deployment, it executes the first ten actions before querying the policy again.
Inference uses ten flow-integration steps.

The training recipe~\cite{shukor2025smolvla} uses AdamW with betas $(0.9,0.95)$, $\epsilon=10^{-8}$, weight decay $10^{-10}$, and gradient-norm clipping at ten.
A 1,000-update warmup is followed by cosine decay from $10^{-4}$ to $2.5\times10^{-6}$ over the 100k-update schedule.
Training uses bf16 autocast and \texttt{torch.compile}.

The principal Blocks baseline runs use 40k-update schedules to match ELMO.

\subsection{SmolVLA + FAST}
\label{sec:supp_fast}

\paragraph{Action representation.}
FAST~\cite{pertsch2025fast} replaces the continuous action output with autoregressive prediction of discrete DCT/BPE tokens.

The BPE vocabulary contains 1,024 entries, with one additional EOS token and one padding token, yielding 1,026 token IDs.
The tokenizer is fitted separately on each benchmark's training actions, with minimum BPE frequency two.
The fitted sequence capacities are 51 slots for LIBERO and 61 slots for CALVIN.

\paragraph{Policy training and inference.}
The SmolVLA expert predicts tokens with causal attention and cross-entropy supervision.
Training uses teacher forcing: each output position receives the preceding target token.
Inference uses greedy autoregressive decoding.
The frozen codec reconstructs actions through inverse BPE, inverse quantization, and inverse DCT.

On LIBERO and CALVIN, FAST represents ten actions and executes all ten.
The final Blocks codec represents 17 actions, approximately one second, and uses 107 token slots.
Blocks evaluation executes ten actions per query.

The policy recipe (taken from EDAR~\cite{xu2026edar}) uses a 1k-update warmup and a 100k-update cosine schedule from $1.5\times10^{-4}$ to $5\times10^{-6}$.
Training uses fp32 with compilation.

\subsection{EDAR}
\label{sec:supp_edar}

\paragraph{Two-stage training.}
EDAR first learns an action/effect representation.
The representation model is then frozen, and SmolVLA is trained to predict its latent registers.
The final implementation matches the paper~\cite{xu2026edar}.

\paragraph{Representation architecture.}
The encoder receives a 16-step action sequence and frozen DINOv3-Base features from the current camera images.
It contains eight encoder layers and eight decoder layers, with width 512, eight attention heads, feed-forward expansion four, and RMSNorm.
Four learned registers produce the continuous action representation.
Each register has seven dimensions on LIBERO and CALVIN.

The encoder jointly processes action tokens, registers, and current visual features.
The decoder uses shared attention with separate action and visual feed-forward paths.
Action queries are causal and read the latent and action context.
The visual branch additionally reads current image features and predicts future visual features.
The action decoding path does not require a live observation.

\paragraph{Representation objective.}
The representation objective combines action reconstruction with future-effect prediction:
\begin{equation}
    \mathcal{L}_{\mathrm{codec}} = \mathcal{L}_{\mathrm{act}} + \lambda_{\mathrm{effect}}\mathcal{L}_{\mathrm{effect}}.
\end{equation}
Action reconstruction sums squared errors over action coordinates and averages over valid time steps.
The effect loss computes cosine distance after flattening each view's patch-feature tensor, then averages across views and samples.
We use $\lambda_{\mathrm{effect}}=0.2$ for LIBERO and $0.1$ for CALVIN.

DINOv3-B uses 224-pixel inputs and retains its native $14\times14$ patch grid.
Inputs use ImageNet normalization and bilinear resizing.
Flattening each view for the cosine objective is the implementation's explicit interpretation of the effect-loss reduction.

\paragraph{Representation optimization.}
Codec training follows the paper~\cite{xu2026edar} and uses Adam with betas $(0.9,0.99)$, batch size 64, 100k updates, and cosine learning-rate decay from $10^{-4}$ to $10^{-5}$.
Gradient norm is clipped at one.
The final codec uses all training frames without an additional codec holdout.

\paragraph{Policy optimization and deployment.}
After freezing the codec, the SmolVLA flow expert predicts its four register tokens directly.
The register values are not standardized using fitted latent statistics.
Policy training~\cite{xu2026edar} uses batch size 64 and a 100k-update schedule with 1k warmup and cosine decay from $1.5\times10^{-4}$ to $5\times10^{-6}$.
Inference retains the stock ten-step flow sampler.
The decoded action horizon is 16, of which eight actions are executed before replanning.

On LIBERO and CALVIN, the principal recipe uses 100k codec updates and 100k policy updates.
We train one codec for all three policy seeds. 

On Blocks, the codec is trained for 20k updates, followed by the
principal 40k-update policy runs.
The main LIBERO/CALVIN policies use bf16 and compilation.
Blocks use fp32 and compilation.

\subsection{MINT-30M}
\label{sec:supp_mint}

\paragraph{Policy architecture.}
MINT~\cite{huang2026mimic} follows its own implementation rather than the shared SmolVLA backbone.
Our implementation matches the MINT-30M configuration specified in the paper rather than the released MINT-Light.

Visual features concatenate outputs from frozen DINOv2 ViT-L/14 with register tokens and frozen SigLIP SO400M/14, using 224-pixel inputs.
Language is encoded by frozen BERT-base.
Its pooled representation conditions the policy through FiLM/AdaLN modulation, without language-token cross-attention.

The trainable policy has eight layers, hidden width 256, twelve attention heads with head dimension 64, and feed-forward width 1,024.

\paragraph{SDAT action tokenizer.}
MINT represents actions with a multiscale residual vector-quantized tokenizer, SDAT.
The policy predicts coarse-to-fine codes, and a frozen SDAT decoder reconstructs the action window.

\begin{table}[t]
    \centering
    \small
    \begin{tabular}{lcc}
        \toprule
        Setting & LIBERO & CALVIN \\
        \midrule
        Action horizon & 16 & 32 \\
        Scales & $(1,2,4)$ & $(1,2,3,4)$ \\
        Total code positions & 7 & 10 \\
        Codebook size & 512 & 512 \\
        Code dimension & 32 & 32 \\
        \bottomrule
    \end{tabular}
    \caption{SDAT tokenizer configurations.}
    \label{tab:supp_mint_tokenizer}
\end{table}

Tokenizers are retrained on each benchmark's training actions.
The final LIBERO baseline does not use the released LIBERO-130 tokenizer.
Tokenizer training uses AdamW, batch size 1,024, learning rate $3\times10^{-5}$, weight decay 0.01, betas $(0.9,0.95)$, and parameter EMA 0.99.
It retains the implementation's reconstruction, vector-quantization, and DCT objectives.
LIBERO trained for 25,300 updates and CALVIN for 26,700 tokenizer updates.
Both reached similar held-out validation MSE.

\paragraph{Policy optimization.}
Policy training uses batch size 128 and 30k updates.
AdamW uses learning rate $3\times10^{-4}$, betas $(0.9,0.95)$, and weight decay 0.01.
The schedule uses 1k warmup followed by cosine decay to $9\times10^{-5}$.
Feed-forward dropout is 0.05, and policy parameter EMA is 0.999.

\paragraph{Deployment.}
Inference uses EMA parameters, top-three code sampling, and intent-weighted temporal ensembling of overlapping action chunks.
The ensemble combines intent similarity, prediction confidence, and temporal decay.
The policy executes four actions per query.

\paragraph{Image augmentation.}
The published version uses image augmentations. 
Since other baselines and ELMO do not use image augmentations, we train a version without to be more directly comparable.
The augmented configuration applies up to two image transformations selected from brightness, contrast, saturation, hue, sharpness, and affine transformations.
These are reported separately, since augmentation substantially affects LIBERO-Plus robustness.

\paragraph{Bimanual Blocks adaptation.}
For Blocks, the model receives three cameras and uses a shared per-arm SDAT tokenizer over seven-dimensional action streams.
The tokenizer uses horizon 32, scales $(1,2,3,5,8)$, and a $1{,}024\times32$ codebook.
Codes for both arms are predicted at each scale.
The principal Blocks policies train for 40k updates without image augmentation because color is essential for instruction following.

\subsection{Detailed Simulation Results}
\label{sec:supp-simulation}

\paragraph{LIBERO-Plus Perturbation Breakdown.}
For LIBERO-Plus~\cite{fei2025liberoplus}, we evaluate the LIBERO-trained checkpoints zero-shot across all seven perturbation families: object layout, camera viewpoint, robot initial state, language paraphrasing, lighting, background texture, and sensor noise.
Table~\ref{tab:plus_perturbation} reports the per-perturbation success rates alongside the overall average across all evaluated episodes.

\begin{table*}[t]
  \centering
  \caption{\textbf{LIBERO-Plus success rate (\%) by perturbation type.} Mean $\pm$ std error across $3$ seeds.}
  \label{tab:plus_perturbation}
  \resizebox{\textwidth}{!}{%
  \begin{tabular}{lcccccccc}
    \toprule
    Method & Camera & Robot Init. & Language & Light & Background & Noise & Layout & Total \\
    \midrule
    SmolVLA~\cite{shukor2025smolvla} & $53.3 \pm 3.2$ & $26.7 \pm 1.3$ & $11.9 \pm 0.3$ & $81.3 \pm 1.6$ & $76.6 \pm 0.8$ & $55.9 \pm 0.9$ & $61.6 \pm 0.9$ & $50.2 \pm 0.8$ \\
    SmolVLA + FAST~\cite{pertsch2025fast} & $54.5 \pm 1.9$ & $24.0 \pm 0.9$ & $15.9 \pm 0.7$ & $78.5 \pm 2.4$ & $71.2 \pm 1.3$ & $57.7 \pm 2.0$ & $62.1 \pm 0.6$ & $50.1 \pm 0.3$ \\
    SmolVLA + EDAR~\cite{xu2026edar} & $55.3 \pm 0.5$ & $28.7 \pm 0.2$ & $19.2 \pm 0.7$ & $\mathbf{86.7 \pm 1.1}$ & $80.4 \pm 1.4$ & $62.2 \pm 2.0$ & $64.2 \pm 0.8$ & $54.4 \pm 0.2$ \\
    \midrule
    MINT-30M~\cite{huang2026mimic} & $58.7 \pm 1.3$ & $\mathbf{49.9 \pm 1.8}$ & $20.3 \pm 0.7$ & $86.3 \pm 1.1$ & $\mathbf{81.3 \pm 2.2}$ & $72.9 \pm 2.2$ & $\mathbf{70.1 \pm 0.9}$ & $61.0 \pm 1.3$ \\
    MINT-30M (w/o image aug.) & $35.0 \pm 2.8$ & $45.5 \pm 1.6$ & $18.8 \pm 1.2$ & $75.5 \pm 0.9$ & $64.9 \pm 2.0$ & $46.7 \pm 2.7$ & $61.3 \pm 1.0$ & $47.8 \pm 1.3$ \\
    \midrule
    \textbf{ELMO (Ours)} & $\mathbf{88.3 \pm 1.0}$ & $36.7 \pm 1.4$ & $\mathbf{46.0 \pm 1.4}$ & $72.5 \pm 2.1$ & $47.5 \pm 2.8$ & $\mathbf{88.3 \pm 1.1}$ & $64.2 \pm 0.2$ & $\mathbf{64.0 \pm 0.8}$ \\
    \bottomrule
  \end{tabular}}
\end{table*}

\paragraph{CALVIN Subtask Completion.}
On the CALVIN ($\text{ABCD}\rightarrow\text{D}$) benchmark~\cite{mees2021calvin}, policies execute 500 chains of five sequential instructions in the held-out environment D.
Table~\ref{tab:calvin-length} reports the success rates for completing at least $k \in \{1, 2, 3, 4, 5\}$ consecutive subtasks alongside the mean chain length.

\begin{table}[t]
  \centering
  \caption{\textbf{CALVIN (ABCD$\rightarrow$D) sequence completion.} Columns $1\text{--}5$ report the percentage of chains completing at least $k$ consecutive subtasks, and Length is the sum of these five completion probabilities. Mean over three seeds.}
  \label{tab:calvin-length}
  \resizebox{\columnwidth}{!}{%
  \begin{tabular}{lcccccc}
    \toprule
    Method & 1 & 2 & 3 & 4 & 5 & Length $\uparrow$ \\
    \midrule
    SmolVLA~\cite{shukor2025smolvla} & 90.1 & 74.3 & 59.6 & 47.3 & 35.8 & 3.07 \\
    SmolVLA + FAST~\cite{pertsch2025fast} & 89.9 & 77.5 & 65.4 & 55.1 & 44.9 & 3.33 \\
    SmolVLA + EDAR~\cite{xu2026edar} & 90.5 & 75.9 & 65.0 & 56.1 & 45.1 & 3.32 \\
    MINT-30M~\cite{huang2026mimic} & 86.1 & 68.8 & 51.6 & 36.5 & 25.4 & 2.68 \\
    \textbf{ELMO (Ours)} & \textbf{97.0} & \textbf{90.5} & \textbf{82.0} & \textbf{73.3} & \textbf{64.8} & \textbf{4.08} \\
    \bottomrule
  \end{tabular}}
\end{table}

\section{Blocks Benchmark Details}
\label{sec:supp-blocks}

\subsection{Motivation, Scene, and Observation Space}
\label{sec:supp_blocks_scene}

Blocks is a language-conditioned tabletop manipulation benchmark built in MuJoCo~\cite{todorov2012mujoco} on the bimanual ALOHA platform~\cite{zhao2024aloha} (\texttt{google-deepmind/aloha\_sim}) with two opposing 6-DoF ViperX 300s arms, one overhead RGB camera, and two wrist-mounted RGB cameras.
It evaluates object selection, spatial generalization, bimanual manipulation, and instruction following while avoiding the spatial/proximity shortcuts, missing language conditioning, or complex real-world transfer barriers of prior benchmarks (Table~\ref{tab:benchmark_comparison}).
Every scene contains the same five persistent $5$\,cm cubes ($0.05$\,kg each) colored red, green, blue, yellow, and purple, ensuring that object identity cannot be inferred merely from which objects are present in the scene.

\begin{table}[t]
    \centering
    \small
    \setlength{\tabcolsep}{4pt}
    \caption{\textbf{Comparison of manipulation benchmarks.} Existing benchmarks either permit spatial/proximity shortcuts~\cite{jiang2026we} and thus lack fully random object locations, lack language conditioning, or involve complex setups that hinder direct sim-to-real transfer.}
    \label{tab:benchmark_comparison}
    \begin{tabular}{lccc}
        \toprule
        \textbf{Benchmark} & \textbf{\begin{tabular}[c]{@{}c@{}}Random \\ Locations\end{tabular}} & \textbf{\begin{tabular}[c]{@{}c@{}}Language\\ Instruction\end{tabular}} & \textbf{\begin{tabular}[c]{@{}c@{}}Easy Setup\\ Transfer To Real\end{tabular}} \\
        \midrule
        LIBERO~\cite{liu2023libero} & \xmark & \cmark & \xmark \\
        CALVIN~\cite{mees2021calvin} & \xmark & \cmark & \xmark \\
        RoboTwin 2.0~\cite{chen2026robotwin} & \xmark & \cmark & \xmark \\
        Colosseum V2~\cite{morgan2026colosseumv2} & \xmark & \cmark & \xmark \\
        OCBench~\cite{park2026behavioral} & \cmark & \xmark & \xmark \\
        InstructMove~\cite{zhao2026instructmove} & \cmark & \cmark & \xmark \\
        \midrule
        \textbf{Blocks (Ours)} & \cmark & \cmark & \cmark \\
        \bottomrule
    \end{tabular}
\end{table}

\begin{table}[t]
    \centering
    \small
    \caption{\textbf{Standard Blocks simulation settings.}}
    \label{tab:supp_blocks_physics}
    \begin{tabular}{ll}
        \toprule
        Component & Setting \\
        \midrule
        Arm-base positions & $x=\pm0.469$\,m \\
        Cube edge length & $0.05$\,m \\
        Cube mass & $0.05$\,kg \\
        Tabletop height & $0.74$\,m \\
        Physics timestep & $0.002$\,s \\
        Physics substeps per action & 30 \\
        Control interval & $0.06$\,s \\
        Control frequency & Approximately $16.7$\,Hz \\
        Cube/table sliding friction & 0.5 \\
        Arm gravity compensation & Enabled \\
        Initial arm pose & Neutral, grippers open \\
        \bottomrule
    \end{tabular}
\end{table}

Standard simulation parameters are summarized in Table~\ref{tab:supp_blocks_physics}.
The tabletop workspace spans
\begin{equation}
    x\in[-0.32,0.32]\ \mathrm{m},
    \qquad
    y\in[-0.22,0.18]\ \mathrm{m}
\end{equation}
at height $z=0.74$\,m.
Cubes are initialized with pairwise center-to-center separations of at least $10$\,cm and initial yaw angles sampled uniformly from $[-\pi/4, +\pi/4]$.
The simulator steps physics at $2$\,ms ($0.002$\,s) and applies control actions every 30 physics substeps, yielding a $60$\,ms control interval ($16.7$\,Hz).
The simulator accepts 14-dimensional commands:
\begin{equation}
    q =
    \left[
    q^L_{1:6},g^L,q^R_{1:6},g^R
    \right],
\end{equation}
where $g=1$ denotes an open gripper and $g=0$ a closed gripper.
Policies receive only RGB images, state, and the current language instruction.

\paragraph{Cameras and visual preprocessing.}
The overhead camera renders at $1280\times720$; the simulator crops the rectangle from pixel $(330,291)$ to $(950,663)$ and resizes it to $480\times288$.
The left and right wrist cameras each render at $384\times216$.
ELMO feature construction applies an additional overhead-image crop from $(48,0)$ to $(432,264)$, followed by bilinear resizing to $512\times512$.

\subsection{Tasks, Sampling, and Generalization Splits}
\label{sec:supp_blocks_tasks}

\paragraph{Task families and sampling.}
Language instructions are instantiated from four templates across three task families:
\begin{itemize}
    \item \texttt{stack the A block on the B block};
    \item \texttt{move the A block to the SIDE of the table};
    \item \texttt{move the A and B blocks to the SIDE of the table};
    \item \texttt{push the A block to the DIRECTION}.
\end{itemize}
For move tasks, destination regions on the tabletop are defined by
\begin{align}
    \text{left}:   &\quad x<-0.15\ \mathrm{m}, &
    \text{right}:  &\quad x> 0.15\ \mathrm{m},\\
    \text{top}:    &\quad y> 0.06\ \mathrm{m}, &
    \text{bottom}: &\quad y<-0.10\ \mathrm{m},
\end{align}
which may overlap at the corners; expert placement targets lie at least $2$\,cm inside the requested boundary.

To prevent frequent argument rejection from skewing the task-kind distribution, the task sampler first randomizes the task-kind order and then attempts up to 60 feasible argument draws for the selected kind before advancing to another kind.
For move tasks, a second block is proposed with probability $0.4$ when available.
Blocks covered by another block are excluded from the movable set, and a move instruction is rejected if any named block already lies inside the requested region.
Push tasks require the target block to rest on the table with a clear, reachable motion corridor.

\paragraph{Color--position generalization split.}
To evaluate compositional spatial generalization, the workspace is partitioned into a $4\times2$ grid defined by $x$-boundaries $(-0.32,-0.16,0,0.16,0.32)$\,m and $y$-boundaries $(-0.22,-0.02,0.18)$\,m.
Using zero-based grid indices, each cube color is excluded from resting or sliding at tabletop height inside one assigned cell during training: red $(0,1)$, green $(3,0)$, blue $(0,0)$, yellow $(1,0)$, and purple $(3,1)$.
Carrying a grasped cube in free space above its held-out cell remains permitted.
All evaluation seeds are strictly disjoint from training seeds.
For position out-of-distribution (position-OOD) evaluation, a designated block is initialized inside its held-out cell and the instruction is forced to name that block.

\subsection{Scripted Demonstration Expert and Dataset}
\label{sec:supp_blocks_expert}

Training demonstrations are collected by a scripted inverse-kinematics expert following Cartesian waypoints.
To induce trajectory diversity, each task receives a speed multiplier sampled from $[0.85,1.15]$, grasp positions receive up to $\pm2$\,mm XY jitter, and approach heights receive up to $\pm1$\,cm jitter.

\paragraph{Arm parking and settling.}
Before an active arm enters the shared central workspace, the inactive arm returns to its neutral parked pose.
The active arm decelerates during final pre-contact descent.
After each motion segment, the expert holds its target until the measured grasp center settles within $4$\,mm of the commanded center, for at most 10 additional control steps.

\paragraph{Cross-table relay.}
When a task requires transporting a cube across single-arm reach limits, the expert executes a bimanual handoff via a central relay zone.
The source arm places the block at a candidate relay location ($x\in\{-3, 0, 3\}$\,cm combined with candidate $y$ coordinates and small random offsets) and retreats to its parked pose before the receiving arm retrieves the block.

\paragraph{Pushing.}
Pushes use a half-open gripper and low contact height to prevent tipping.
Long pushes are executed via at most four short corrective strokes (each at most $12$\,cm) aligned with the cube's live position.

\paragraph{Pose-recovery demonstrations.}
To expose the policy to off-nominal states and the corrections that follow them, we generate an additional set of demonstrations in which the expert acts on a deliberately wrong estimate of its target.
For each pick, place, or push approach, with probability $0.7$ the expert's estimate of the target pose is offset by $1.5$--$4.5$\,cm in a uniformly random horizontal direction, with an additional yaw error of up to $\pm0.15$\,rad.
The expert plans its approach on this estimate: it moves above the wrong position and descends to $10$\,cm above the target while still centred on it.
The estimate then converges linearly with height to the true pose while the gripper continues to descend slowly, so the arm visibly corrects towards the cube before the normal grasp, place, or push begins.
Every frame is labelled with its phase. Actions of the approach towards the wrong estimate are masked from the action loss, so the policy is never trained to imitate the error itself.

\paragraph{Filtering and dataset composition.}
During data generation, a demonstration episode is retained only if the expert completes the task without disturbing uninstructed cubes, violating spatial holdouts, or incurring inter-arm collisions.
We generated $10{,}000$ clean tasks and $2{,}500$ pose-recovery tasks. 
After filtering for collisions and invalid block locations, the collection consists of $8{,}441$ clean tasks and $2{,}097$ pose-recovery tasks, in which the pick, place, or push point is randomly perturbed so the expert must execute corrective adjustments before manipulating the block.

\subsection{Scoring and Evaluation}
\label{sec:supp_blocks_eval}

\paragraph{Goal predicates and failure conditions.}
Let $p_b$ denote the 3D center of block $b$ and $p_b^0$ its position at the start of the task.
A task succeeds only when for ten consecutive frames its geometric goal predicate is satisfied and both grippers are open:
\begin{itemize}
    \item \textbf{Stacking ($A$ on $B$):} The instructed upper cube $A$ must lie within $2.5$\,cm horizontally of the lower cube $B$ with a vertical center separation of $5.0\pm1.2$\,cm:
    \begin{align}
        |p_{A,x}-p_{B,x}| &< 0.025\ \mathrm{m},\\
        |p_{A,y}-p_{B,y}| &< 0.025\ \mathrm{m},\\
        |p_{A,z}-p_{B,z}-0.05| &< 0.012\ \mathrm{m}.
    \end{align}
    \item \textbf{Moving:} Every instructed cube must rest within $1.2$\,cm of tabletop height inside the target region: left ($x<-0.15$\,m), right ($x>0.15$\,m), top ($y>0.06$\,m), or bottom ($y<-0.10$\,m).
    \item \textbf{Pushing:} Let $u$ be the commanded cardinal unit vector in the tabletop plane, $u_\perp$ its perpendicular unit vector, and $d=p_{A,xy}-p^0_{A,xy}$. The instructed cube $A$ must translate by at least $7$\,cm along $u$, deviate by at most $5$\,cm laterally, and remain within $1.2$\,cm of tabletop resting height:
    \begin{align}
        d^\top u &\geq 0.07\ \mathrm{m}, &
        |d^\top u_\perp| &\leq 0.05\ \mathrm{m}.
    \end{align}
\end{itemize}
If any uninstructed cube $b\notin\mathcal{B}_{\mathrm{named}}$ (where both blocks in a stacking instruction count as named) moves by more than $3$\,cm in 3D Euclidean distance from its initial position,
\begin{equation}
    \exists b\notin\mathcal{B}_{\mathrm{named}}:
    \|p_b-p_b^0\|_2>0.03\ \mathrm{m},
\end{equation}
the rollout terminates immediately as a wrong-object failure.
This wrong-block check takes precedence over goal success.
Any task exceeding $450$ control steps times out.

\paragraph{Evaluation tracks and metrics.}
Standard evaluation covers 500 in-distribution (ID) tasks sampled from the training initial-position distribution and 250 position-OOD tasks where the named target starts in its held-out color--position cell.

\subsection{Task Breakdown and Diagnostic Comparisons}
\label{sec:supp_blocks_diagnostics}

Table~\ref{tab:blocks-breakdown} disaggregates Blocks performance by task family (Pushing, Moving, and Stacking) on a shared evaluation manifest alongside episode-level failure modes (wrong-object disturbance and timeout rates).

\begin{table*}[t]
  \centering
  \caption{\textbf{Blocks task-family and failure-mode breakdown.} Results on the in-distribution (ID) and out-of-distribution (OOD) tasks, mean over three seeds.
  Push, Move and Stack report task success rates (\%). Wrong object and Timeout report the percentages of all evaluation rollouts terminating with each failure condition.}
  \label{tab:blocks-breakdown}
  \resizebox{\textwidth}{!}{%
  \begin{tabular}{lccccc@{\hskip 12pt}ccccc}
    \toprule
    & \multicolumn{5}{c}{ID} & \multicolumn{5}{c}{OOD} \\
    \cmidrule(lr){2-6} \cmidrule(lr){7-11}
    Method & Push $\uparrow$ & Move $\uparrow$ & Stack $\uparrow$ & Wrong obj. $\downarrow$ & Timeout $\downarrow$
      & Push $\uparrow$ & Move $\uparrow$ & Stack $\uparrow$ & Wrong obj. $\downarrow$ & Timeout $\downarrow$ \\
    \midrule
    SmolVLA~\cite{shukor2025smolvla} & 12.9 & 0.2 & 0.0 & 23.7 & 71.4 & 2.2 & 0.0 & 0.0 & 24.1 & 75.2 \\
    SmolVLA + FAST~\cite{pertsch2025fast} & 8.1 & 0.4 & 0.2 & 33.5 & 63.3 & 1.8 & 0.0 & 0.0 & 37.2 & 62.3 \\
    SmolVLA + EDAR~\cite{xu2026edar} & 12.5 & 0.2 & 0.0 & 24.7 & 70.5 & 5.7 & 0.7 & 0.0 & 20.9 & 77.1 \\
    MINT-30M~\cite{huang2026mimic} & 5.4 & 0.0 & 0.0 & 15.4 & 82.6 & 3.5 & 0.0 & 0.0 & 14.8 & 84.1 \\
    \textbf{ELMO (Ours)} & \textbf{83.0} & \textbf{17.2} & \textbf{7.2} & \textbf{10.5} & \textbf{50.9}
      & \textbf{74.6} & \textbf{15.2} & \textbf{10.4} & \textbf{13.1} & \textbf{54.8} \\
    \bottomrule
  \end{tabular}}
\end{table*}

\section{Real-World Evaluation Details}
\label{sec:supp-real}

\paragraph{Evaluation Setup.}
We conduct our real-world evaluation on a standard bimanual ALOHA setup, consisting of two parallel-jaw follower arms—each equipped with a wrist-mounted camera—alongside an overhead scene camera and a lower desk camera (unused). The setup also includes two leader arms for collecting training demonstrations via teleoperation. This physical environment does not identically replicate the simulated Blocks environment, differing primarily in joint dynamics (e.g., controller gains), camera extrinsics (e.g., wrist camera placement and viewpoints), and visual appearance (e.g., lighting and tabletop textures). The blocks are 3D-printed from PLA with a side length of $5\,\mathrm{cm}$ and painted to approximately match the colors used in simulation.

\paragraph{Training Procedure.}
For both the SmolVLA baseline and our model (ELMO), we initialize from the best simulation checkpoints and fine-tune on real-world teleoperated demonstrations for $10\mathrm{k}$ steps, reinitializing the optimizer and learning rate schedule. We select $10\mathrm{k}$ steps based on the convergence of the training and validation curves. Rather than co-training on combined simulation and real-world data, we fine-tune exclusively on real-world data, as this yields stronger performance for both models—likely due to the substantial sim-to-real gap between the simulated and physical setups.

We collect $30$ demonstrations for each combination of task family (push and move), block color, and direction, totaling $1{,}200$ demonstrations across $40$ distinct task combinations. Before each demonstration, the blocks are reset to randomized positions. Additionally, we crop the wrist and scene camera feeds to better align their fields of view with the simulation environment.

\paragraph{Inference Procedure.}
Both models follow the same inference procedure used in simulation. 
Because deploying the policies directly on the physical robot causes excessive jitter, we smooth action execution using temporal ensembling from ACT~\cite{zhao2023act}.
Specifically, at each control step, we compute the executed action as an exponentially weighted average over all overlapping action predictions for the current timestep, weighting the $i$-th prediction by $w_i = \exp(-m \cdot i)$ with $m=0.05$. While querying the policy at every timestep ($k=1$) maximizes smoothing, the higher computational overhead of SmolVLA prevents real-time execution at that rate. Thus, we query SmolVLA every $k=10$ steps, whereas ELMO is queried at every step ($k=1$).

\paragraph{Evaluation Procedure.}
Before each trial, we place the blocks at randomized positions on the table, and an automated script samples the target color and direction. For the push task, a trial is marked successful if the robot executes a clear pushing motion (rather than grasping) that moves the target block along an approximately straight path in the specified direction. For the move task, success requires picking up the target block and placing it in the location corresponding to the specified direction. Each policy is allotted $20\,\mathrm{s}$ per trial; timeouts and excessive contact with non-target blocks are counted as failures.

\section{Additional Latent Space Analysis Details}
To compare what each method's latent encodes, we encode the same demonstration chunks with every model and fit linear probes and t-SNE embeddings on the result.
For each benchmark we sample 6{,}000 chunk start points from the training demonstrations, on a stride of 8 steps and only where a full action horizon remains, using a fixed random seed.
Each chunk is labelled with the instruction, the task group, the target object, whether the gripper opens or closes within the next 16 steps, the episode progress, the end-effector position at the start and 16 steps later, the end-effector displacement over those 16 steps, the commanded gripper state, and (CALVIN, Blocks) the target block position and the positions of all blocks.
Every model encodes the chunk with its own frozen representation, using the inputs it receives during policy training:
\begin{itemize}
  \item \textbf{MINT (all scales):} MINT's policy predicts the tokens of a frozen multi-scale VQ-VAE over action chunks. We take this VQ-VAE from the final MINT policy and tokenise each chunk's ground-truth actions over its native horizon. The latent is the quantised representation $\hat f$ the VQ-VAE decoder reads (the sum of the upsampled codebook embeddings over all scales that the MINT policy learns to predict). For bimanual tasks the two arms are tokenised as separate streams after per-dimension quantile normalisation, and their latents are concatenated.
  \item \textbf{MINT (scale 1):} the codebook embedding of the single coarsest-scale token of the same tokenisation (per arm), which is where the MINT policy places its intention.
  \item \textbf{EDAR:} EDAR's policy is trained to predict the latent code of a frozen action autoencoder, which encodes a chunk of 16 actions together with the camera images at the chunk's first step into a small set of register tokens.
  We take this autoencoder from the final EDAR policy checkpoint and encode each demonstration chunk with it, giving it the ground-truth actions (normalised with the autoencoder's own statistics) and the camera images at the chunk's first step.
  The latent is the encoder's register tokens, flattened into a single vector.
  \item \textbf{ELMO:} the latent plan vector the decoder receives at the chunk's first step, taken from a teacher-forced forward pass over the whole demonstration.
\end{itemize}

\paragraph{Probes.}
All latents are standardised using statistics from the training folds. We fit logistic regression (L2, $C{=}1$) for categorical labels and ridge regression ($\alpha{=}10$) for continuous ones, with 5-fold cross-validation grouped by episode so that no episode appears in both training and test folds.
Categorical probes are scored as accuracy above the majority-class rate, $(\mathrm{acc}-p_{\mathrm{maj}})/(1-p_{\mathrm{maj}})$, and continuous probes as $R^2$ averaged over output dimensions; negative scores are shown as 0 in the radar charts.
As a reference we probe the raw $16$-step action chunk.

\paragraph{t-SNE.}
Each latent is standardised and, if its dimension $d$ exceeds $50$, projected onto its first $50$ principal components (PCA to $\min(50, d)$ dimensions). The result is embedded in 2D with t-SNE (perplexity 30, PCA initialisation). Linear probes are fit on the full standardised latents, without PCA.

\clearpage

\twocolumn[{
    \section{Additional Figures}
    \label{sec:supp-figs}
    \begin{center}
        \includegraphics[trim={0.5cm 8cm 0.5cm 0cm},clip,width=0.95\textwidth]{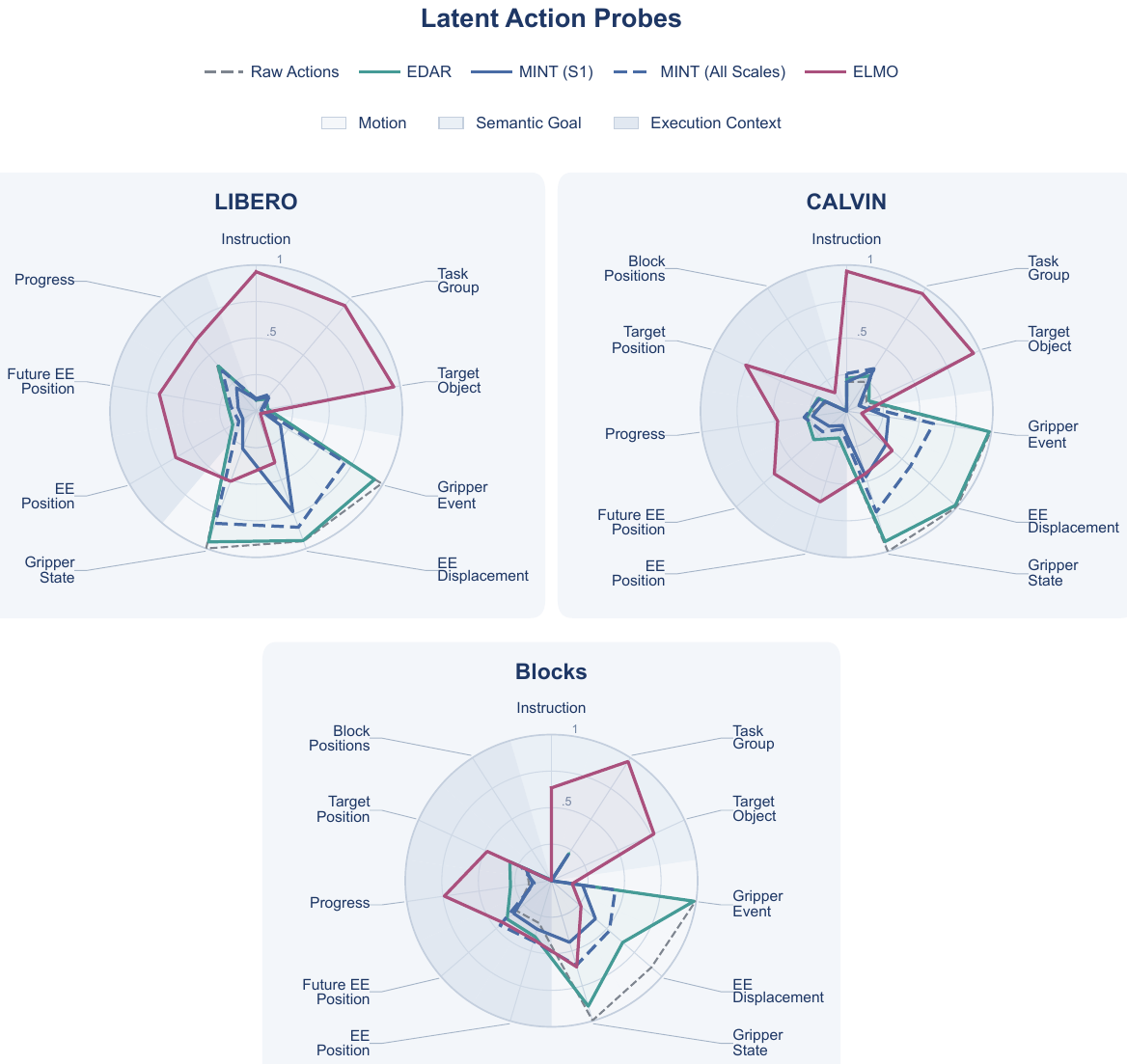}
        \captionof{figure}{Extended linear probe performance on all benchmarks.}
        \label{fig:supp-radar}
    \end{center}
}]

\begin{figure*}[t]
    \centering
    \includegraphics[trim={0.5cm 10.8cm 0.5cm 0cm},clip,width=\linewidth]{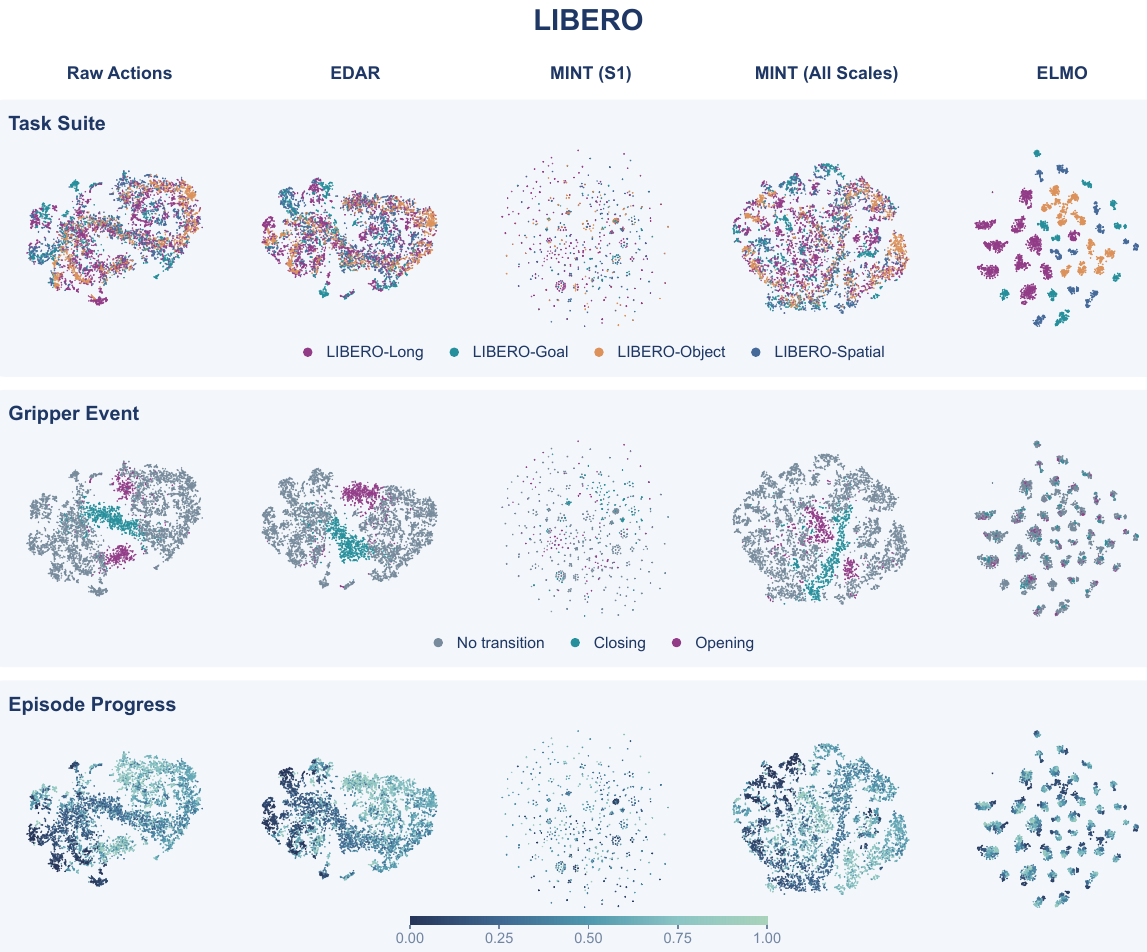}
    \caption{Additional t-SNE plots on the LIBERO benchmark.}
    \label{fig:supp-tsne-libero}
\end{figure*}

\begin{figure*}[t]
    \centering
    \includegraphics[trim={0.5cm 10.8cm 0.5cm 0cm},clip,width=\linewidth]{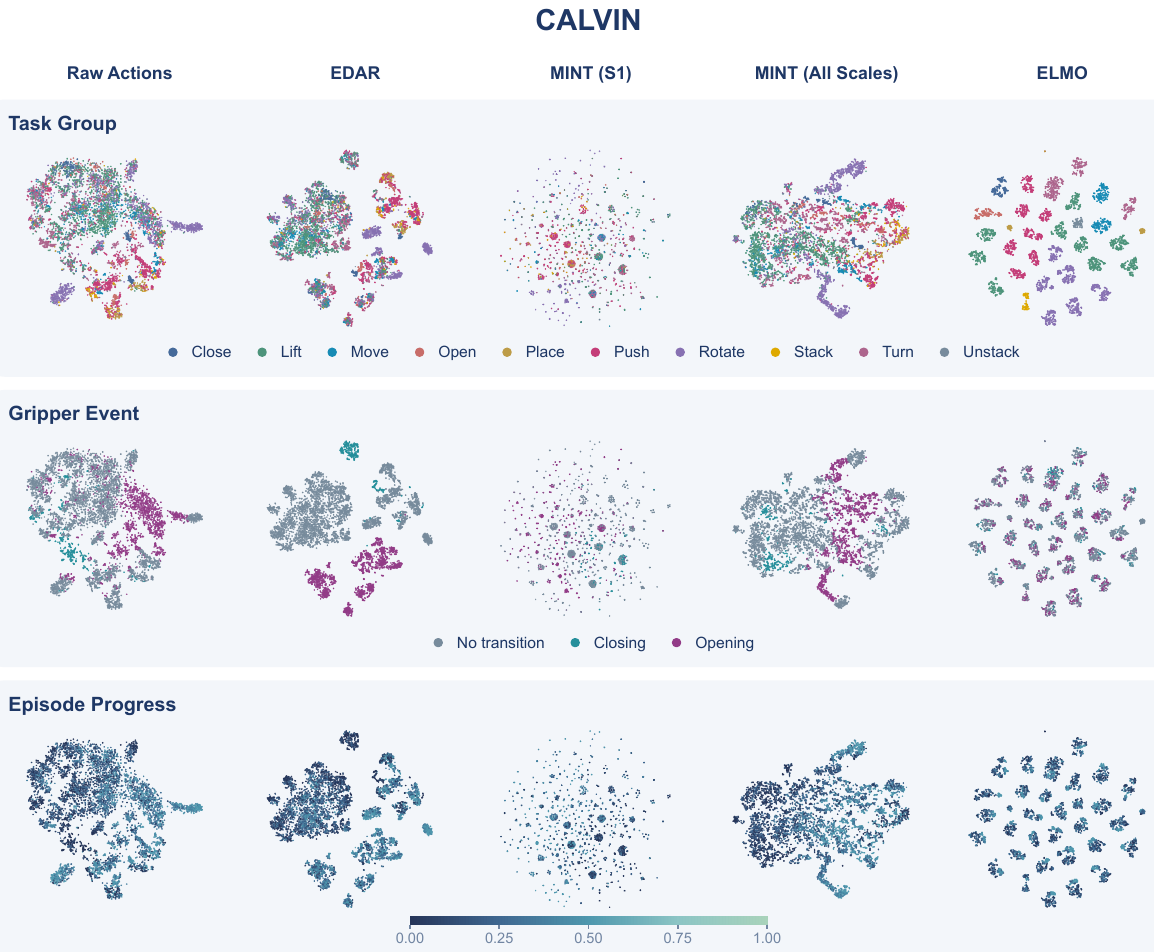}
    \caption{Additional t-SNE plots on the CALVIN benchmark.}
    \label{fig:supp-tsne-calvin}
\end{figure*}

\begin{figure*}[t]
    \centering
    \includegraphics[trim={0.5cm 15.8cm 0.5cm 0cm},clip,width=\linewidth]{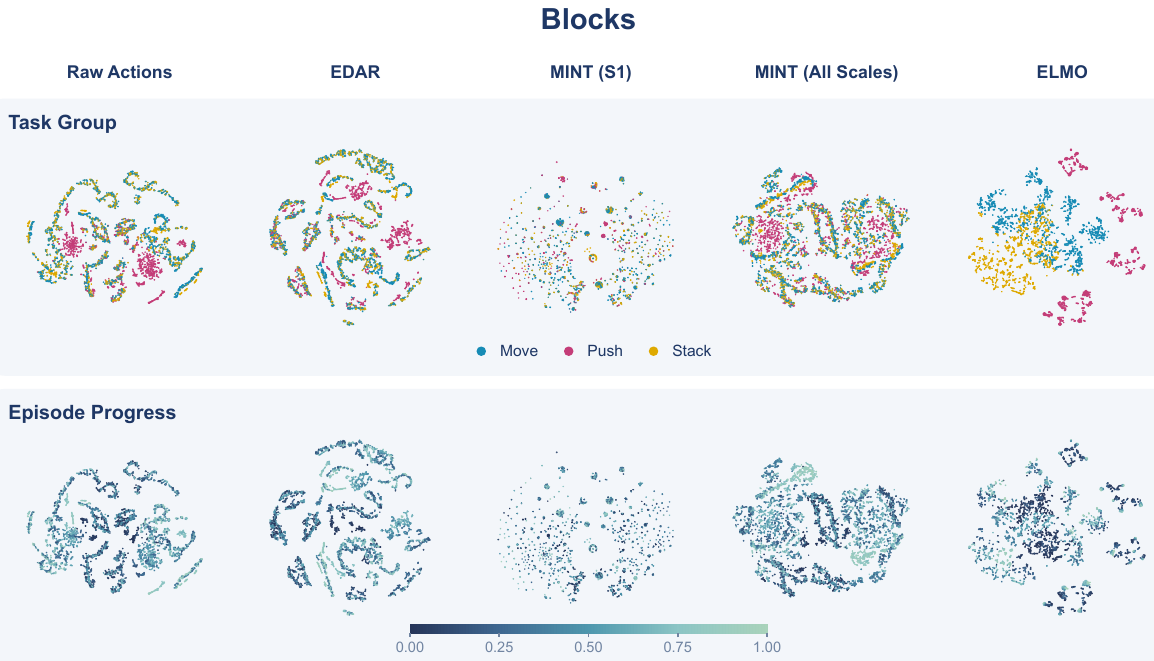}
    \caption{Additional t-SNE plots on the Blocks benchmark.}
    \label{fig:supp-tsne-blocks}
\end{figure*}

\clearpage

\end{document}